\documentclass[11pt,a4paper]{article}
\usepackage{authblk}
\usepackage{graphicx}
\usepackage{amsmath}
\usepackage{amssymb}
\usepackage{algorithm}
\usepackage{algorithmic}
\usepackage{amsthm}
\usepackage{epstopdf}
\usepackage{caption}
\usepackage{url}

\usepackage{natbib}
\usepackage{cases}
\usepackage{bm}
\usepackage{booktabs}
\usepackage{subfig}
\usepackage{multicol} 
\usepackage{multirow}
\usepackage{enumitem}

\newtheorem{proposition}{Proposition}

\theoremstyle{definition}

\title{ Variational Policy Propagation for Multi-agent Reinforcement Learning  }
\date{}
\author[1]{Chao Qu \footnote{Chao Qu and Hui Li are equally contributed. chaoqu.technion@gmail.com}}
\author[1]{Hui Li *}
\author[2]{Chang Liu}
\author[1]{Junwu Xiong}
\author[1]{James Zhang}
\author[1]{Wei Chu}
\author[1]{Weiqiang Wang}
\author[1]{Yuan Qi}
\author[1,3]{Le Song}
\affil[1]{Ant Financial Services Group}
\affil[2]{Shanghai Jiao Tong University}
\affil[3]{College of Computing, Georgia Institute of Technology}

\begin{document}

\maketitle

\begin{abstract}
We propose a \emph{collaborative} multi-agent reinforcement learning algorithm named variational policy propagation (VPP) to learn a \emph{joint} policy through the interactions over agents.  We prove that the joint policy is a Markov Random Field  under some mild conditions, which in turn reduces the policy space effectively. We integrate the variational inference  as special differentiable layers in policy such that the actions can be efficiently sampled from the Markov Random Field and the overall policy is differentiable. We evaluate our algorithm on several large scale challenging tasks and demonstrate that it outperforms previous state-of-the-arts.
\end{abstract}

\section{Introduction}
Collaborative multi-agent reinforcement learning is an important sub-field of the multi-agent reinforcement learning (MARL), where the agents learn to coordinate to achieve joint success. It has wide applications in traffic control, autonomous driving  and smart grid~\citep{kuyer2008multiagent,shalev2016safe}. To learn a coordination, the interactions between agents are indispensable. For instance,  humans can reason about other's behaviors or know other peoples' intentions through communication and then determine an effective coordination plan. However, how to design a mechanism of such interaction rigorously  and at the same time solve the large scale real-world applications is still a challenging problem.

Recently, there is a surge of interest in solving the collaborative MARL problem \citep{foerster2018counterfactual,qu2019value,lowe2017multi}. Among them, joint policy approaches have demonstrated their superiority \citep{rashid2018qmix,sunehag2018value,oliehoek2016concise}.  A straightforward approach is to replace the action in the single-agent reinforcement learning by the joint action $\mathbf{a}=(a_1,a_2,...,a_N)$, while it obviously suffers from the issue of the \emph{exponentially} large action space. Thus several approaches have been proposed to factorize the joint action space to mitigate such issue, which can be roughly grouped into two categories: factorization over policy and factorization over value functions. The former one explicitly assumes that  $\pi(\mathbf{a}|s):=\prod_{i=1}^{N} \pi_i(a_i|s)$, i.e., policies are independent \citep{foerster2018counterfactual,zhang2018fully}.  The counterpart on the value function has a similar spirit but factorizes the joint value function into several utility functions, each just involving the actions of \emph{one agent} \citep{rashid2018qmix,sunehag2018value}.  However, above approaches lack of the interactions between agents, since in their algorithms agent $i$ does not care about the plan of agent $j$.  Indeed, they may suffer from a phenomenon called relative over-generalization in game theory observed by \citet{wei2016lenient,castellini2019representational,palmer2018lenient}. Approaches based on the coordinate graph would effectively prevent such cases, where the value function is factorized as a summation of utility function on pairwise or local joint action \citep{guestrin2002coordinated,bohmer2019deep}. However, they only can be applied in discrete action, small scale game.

Despite the empirical success of the aforementioned work in certain scenarios, it still \emph{lacks} theoretical insight. For instance, which strategy would their algorithm converge to. In this work, we only make a simple yet realistic assumption: the reward function $r_i$ of each agent $i$ just depends on its individual action and the actions of its neighbors (and  state)
\begin{align}
r_i(s,\mathbf{a})=r_i(s,a_i,a_{\mathcal{N}_i}),   
\label{eq:assumption}
\end{align}
where we use $\mathcal{N}_i$ to denote the neighbors of agent $i$, $s$ to denote the global state.
It says the goal or decision of an agent is explicitly influenced by a small subset $\mathcal{N}_i$ of other agents. Note that such an assumption is \emph{reasonable in lots of real scenarios}. For instance, The traffic light at an intersection makes the decision on the phase changing mainly relying on the traffic flow around it and the policies of its neighboring traffic light. 
The main goal of a defender in a soccer game is to tackle the opponent's attacker, while he rarely needs to pay attention to opponent goalkeeper's strategy.  

Based on the assumption in~\eqref{eq:assumption}, we propose a  multi-agent reinforcement learning algorithm inspired by the theory of \emph{variational inference}(VI), where the objective is to maximize the long term reward of the group, i.e., $\sum_{t=0}^{\infty} \sum_{i=1}^{N} \gamma^t r^t_i$ ( plus an entropy term). In particular, we prove that the optimal policy has a Markov Random Field (MRF) form, and therefore reduce the exponentially large joint action space to a much smaller one.  To efficiently sample the action from MRF, we leverage VI, a generic method to approximate complicated distribution \cite{bishop2006pattern} where the objective is to minimize the free energy under certain constraints. However, naively applying VI would block the gradient of the neural network. To address this issue, we device differentiable variational inference layers where the input is (the potential of) MRF while the the output is its corresponding variational approximation (e.g., the mean-field approximation, loopy belief propagation and many others).   The whole multi-agent system can be trained in an end-to-end manner and the policy converge to the optimal solution of the free energy approximation of the joint policy.  We name our algorithm variational policy propagation (VPP) to reveal its connection to VI in nature.

\textbf{Contributions:}  
(1) We propose a theory inspired method named variational policy propagation to solve the joint policy collaborative MARL problem; 
(2) Our method is computationally efficient, which can scale up to \emph{one thousand} agents.
(3) Empirically, it outperforms  state-of-the-art baselines with a wide margin when the number of agents is large;
(4) Our work builds a bridge between MARL and variational inference, which would lead to new algorithms beyond VPP.

\textbf{Notation}:  $s_i^t$ and $a_i^t$ represent the state and action of agent $i$ at time step $t$. The neighbors of  agent $i $  are represented as  $ \mathcal{N}_i$.  We denote $X$ as a random variable with domain $\mathcal{X}$ and refer to instantiations of $X$ by the lower case character $x$. We denote  a density on $\mathcal{X}$ by $p(x)$ and denote the space of all such densities as by $\mathcal{P}$.

\section{Related work}
\vspace{-2mm}
Since there are a plethora of works in MARL, we just list closely related ones here, and defer more discussions to appendix \ref{app:more_related_work}. 

COMA designs a MARL algorithm based on the actor-critic framework with independent actors $\pi_{i}(a_i|s)$, where the joint policy is factorized as $\pi(\mathbf{a}|s)=\prod_{i=1}^{N}\pi_i(a_i|s)$ \citep{foerster2018counterfactual}.   MADDPG considers a MARL with the cooperative or competitive setting, where it creates a critic for each agent \citep{lowe2017multi}. Other similar works may include \citep{de2019multi,wei2018multiagent}. Another way is to factorize the value functions into several utility functions, which may include \citet{sunehag2018value,rashid2018qmix,son2019qtran}.  However these factorized methods suffer from the relative overgeneralization issue  \citep{castellini2019representational,palmer2018lenient}. Generally speaking, it pushes the agents to underestimate a certain action because of the low rewards they receive, while they could get a higher one by perfectly coordinating. A middle ground between the (fully) joint policy and the factorized policy is the coordination graph~\citep{guestrin2002coordinated}, where the value function is factorized as a summation of the utility function on the pairwise action. \citet{bohmer2019deep,castellini2019representational} combine deep learning techniques with the coordination graph. It addresses the issue of relative overgeneralization, but still has two limitations especially in the large scale MARL problem. (1) The max-sum algorithm can just be implemented in the discrete action space since it needs a max-sum operation on the action of $Q$ function. (2) Even in the discrete action case,  \emph{each step} of the $Q$ learning has to do several loops of max-sum operation over the whole graph if there is a cycle in graph. Our algorithm can handle both discrete and continuous action space cases and alleviate the scalability issue by variational inference.

Our work is also related to MARL with  communication. \citet{foerster2016learning} propose an end-to-end method to learn communication protocol.  \citet{chu2020multi} require a strong assumption that the MDP has the spatial-temporal Markov property. They utilize neighbor's action information in a heuristic way and thus it is unclear what the agents are learning (e.g., do they learn the optimal joint policy to maximize the group reward? ). \citet{jiang2020graph} propose DGN which uses GNN to spread the \emph{state} embedding information to neighbors. However each agent still uses an independent Q learning to learn the policy and neglects other agents' plans \citep{bohmer2019deep}. In contrast,  we propose a principled framework, where each agent makes decision considering other agents' plan.  In addition, this framework generates different messages passing algorithm when we change the free energy in the objective function. There are lots of work about reducing the overhead of the communication \citep{ding2020learning,zhang2020succinct}. Our contribution is orthogonal to them, since we focus on a theoretical framework to approximate the joint policy and can directly plug in their methods to reduce the communication cost.

Recently, there are several works \citep{amos2017optnet,donti2017task,donti2021dc3} to integrate the convex optimization problem as individual layers in larger end-to-end trainable deep networks. These work have similar spirit as ours but has a key difference: they consider the optimization over a (convex) function while the objective in VI is a functional. In addition, all these works just apply in supervised learning.

\section{Backgrounds}\label{section:background}
\vspace{-2mm}


\textbf{Probabilistic Reinforcement Learning:} Probabilistic reinforcement learning (PRL)  \citep{levine2018reinforcement} is our building block.  PRL defines the trajectory $\tau$ up to time step $T$ as $\tau=[s^0,a^0, s^1,a^1,...,s^T,a^T,s^{T+1}]$. The probability distribution of the trajectory $\tau$ induced by the optimal policy is defined as $ p(\tau)= [p(s^0)\prod_{t=0}^T p(s^{t+1}|s^t,a^t)]\exp\big(\sum_{t=0}^T r(s^t,a^t) \big).$
While the probability of the trajectory $\tau$ under the policy $\pi(a|s)$ is defined as $\hat{p}(\tau)=p(s^0)\prod_{t=0}^T p(s^{t+1}|s^t,a^t)\pi(a^t|s^t).$
The objective is to minimize the KL divergence between $\hat{p}(\tau)$ and $p(\tau)$ \cite{levine2018reinforcement}. It is equivalent to the  maximum entropy reinforcement learning
$\max_{\pi} J(\pi)=\sum_{t=0}^T \mathbb{E}[r(s^t,a^t)+\mathcal{H}(\pi(a^t|s^t))],$
where it omits the discount factor $\gamma$ and regularizer factor $\alpha$ of the entropy term, since it is easy to incorporate them into the transition and reward respectively.  Such framework subsumes state-of-the-art algorithms such as soft-actor-critic (SAC) \citep{haarnoja2018soft}. In each iteration, SAC optimizes the following  loss function of $Q$,$\pi$, $V$, and respectively.
\begin{flalign*}
&\mathbb{E}_{(s^t,a^t)\sim D} \big[ Q(s^t,a^t)- r(s^t,a^t) - \gamma \mathbb{E}_{s^{t+1}\sim p}[V(s^{t+1})]\big]^2, \\
&\mathbb{E}_{s^t\sim D}\mathbb{E}_{a^t\sim \pi}[\log \pi(a^t|s^t) -Q(s^t,a^t) ]\\
&\mathbb{E}_{s^t\sim D} \big[  V(s^t)-\mathbb{E}_{a^t\sim \pi_\theta} [Q(s^t,a^t)-\log \pi(a^t|s^t)]\big]^2, 
\end{flalign*}
\vspace{-2mm}
where $D$ is the replay buffer.

\textbf{Function Space Embedding of Distribution:}
Essentially, this math tool says that we can map the distribution into a embedding under mild conditions and therefore we can use the embedding in our variational inference layers. 
Embedding of the distribution is a class of non-parametric methods where a probability distribution is represented as an element of a RKHS \citep{smola2007hilbert}. We let $\phi(X)$ be an implicit feature mapping  and $X$ be a random variable with distribution $p(x)$.
Embeddings of $p(x)$  is given by
$\mu_X:= \mathbb{E}_X [\phi(X)]=\int \phi(x)p(x)dx $
where the distribution is mapped to its expected feature map. Under some mild conditions such as embeddings are injective, we can treat the embedding $\mu_X$ of the density $p(x)$ as a sufficient statistic of the density \citep{dai2016discriminative}.  Such property is important since we can reformulate a functional $f:\mathcal{P}\rightarrow \mathbb{R}$ of $p(\cdot)$ using the embedding only, i.e., $f(p(x))=\Tilde{f}(\mu_X)$. It also can be generalized to the operator case. In particular, applying an operator $\mathcal{T}: \mathcal{P}\rightarrow \mathbb{R}^d $ to a density can be equivalently carried out using its embedding $\mathcal{T}\circ p(x)= \Tilde{\mathcal{T}} \circ \mu_X$, where $ \Tilde{\mathcal{T}}: \mathcal{F} \rightarrow \mathbb{R}^d$ is the alternative operator working on the embedding. In practice, $\mu_X$, $\tilde{f}$ and $\tilde{\mathcal{T}}$ have complicated dependence on $\phi$. As such, we  approximate them by neural networks, which is known as the neural embedding approach of distribution \citep{dai2016discriminative}.

\section{Our Method}\label{section:our_method}

\subsection{Setting}
In this section, we present our method VPP for the collaborative multi-agent reinforcement learning.  To begin with, we formally define the problem as a networked MDP \citep{zhang2018fully}. The network is characterized by a graph $\mathcal{G}=(\mathcal{V}, \mathcal{E})$, where each vertex $i\in \mathcal{V}$ represents an agent and the edge $ij\in \mathcal{E}$ means the communication link between agent $i$ and $j$. We say $i$,$j$ are neighbors if they are connected by this edge.  The corresponding networked MDP is characterized by a tuple $( \{\mathcal{S}_{i}\}_{i=1}^{N}, \{\mathcal{A}_{i}\}_{i=1}^{N}, p,\{r_i\}_{i=1}^{N}, \gamma, \mathcal{G})$, where  $N$ is the number of agents,  $S_i$ is the local state of the agent $i$ and $\mathcal{A}_i$ denotes the set of action available to agent $i$. We let $S:=\prod_{i=1}^{N} S_i$ and $\mathcal{A}:= \prod_{i=1}^{N} \mathcal{A}_i$ be the global state and joint action space respectively. At time step $t+1$, the global state $s^{t+1}\in S$ is drawn from the transition   $s^{t+1}\sim p(\cdot| s^{t}, \mathbf{a^t})$, conditioned on the current state $s^t$ and the joint action $\mathbf{a}^t=(a_1^t, a_2^t,...,a_N^t) \in \mathcal{A}.$  Each transition yields a reward $r_i^t=r_i(s^t, \mathbf{a^t})$ for agent $i$ and $\gamma$ is the discount factor.  The aim of our algorithm is to learn a joint policy $ \pi( \mathbf{a^t}| s^t)$ to maximize the overall long term reward plus an entropy term $\mathcal{H}( \pi (\mathbf{a}|s))$ on the joint action $\mathbf{a}$, which encourage the exploration \citep{ziebart2008maximum,haarnoja2018softac}
$$\eta(\pi)=\mathbb{E}{\Large[\sum_{t=0}^{\infty}\gamma^t ( \sum_{i=1}^{N}   r_i^{t}+ \mathcal{H}(\pi(\mathbf{a}^t|s^t)) \Large] },$$ where each agent $i$ can just observe its own state $s_i$ and the message from the neighborhood communication. We denote the neighbors of agent $i$ as $\mathcal{N}_i$ and further assume that the reward $r_i$ depends on the state and the actions of itself and its neighbors, i.e., $r_i(s,\mathbf{a}):=r_i(s,a_i, a_{\mathcal{N}_i})$. 
Such assumption is reasonable in many real scenarios as we discussed in the introduction.

\begin{figure*}
	\centering
	\includegraphics[width=0.98\textwidth]{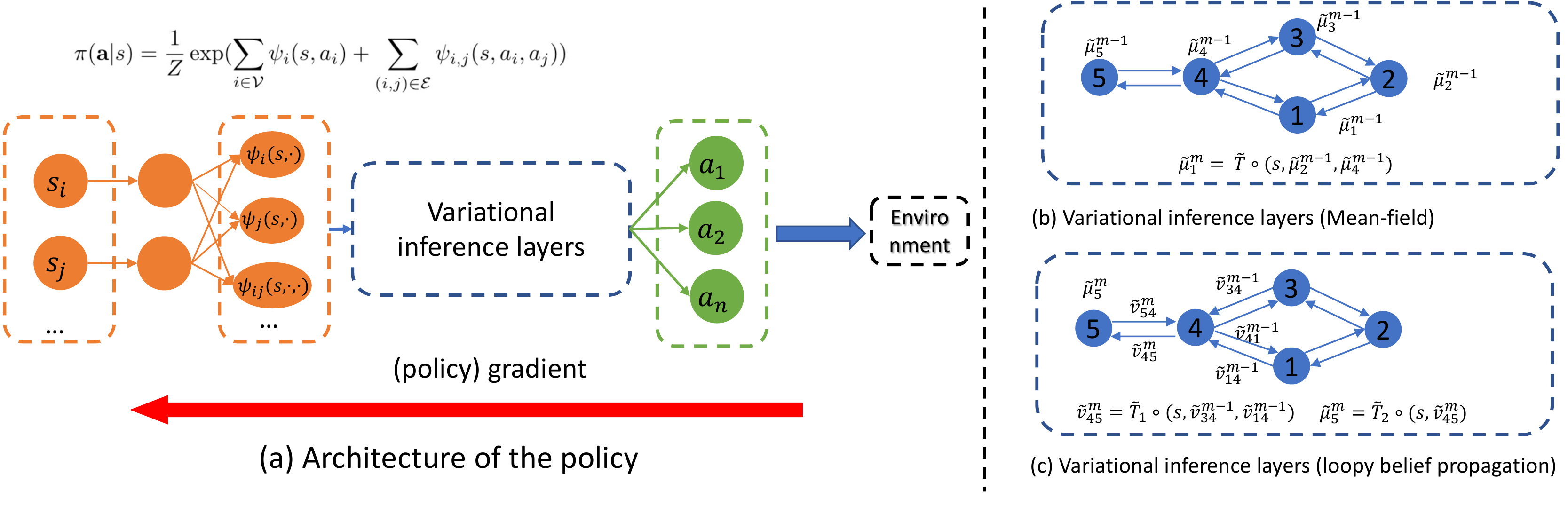}
	\caption{(a) The architecture of the policy. $\psi_i,\psi_j, \psi_{ij}$ are potential functions of MRF. We denote $\psi_i(s,\cdot)$ as a function of $a_i$.  We can use a network represent that, where the input is $s_i$ and the output is a $d$ dimensional vector. We do approximation on $\psi_{ij}$ using similar way but the inputs are the information from node $i$ and $j$.  These embeddings are fed into the variational inference layers,  which are the initial value $ \tilde{\mu}_i^0$ and $\tilde{v}_{ij}^0 $ in  (b) and (c).  Such layers map MRF determined by the potential functions ($\psi_i$ and $\psi_{ij}$ ) to the its variational approximation, and then we sample the action $a_i$ for agent $i$ from this approximation easily (e.g., use the reparametrization trick). Since these layers are differentiable, policy gradient can back-propagate to inputs and the parameters  in variational inference layers and  potential functions can be learned by the gradient. (b)(c) represent two message passing procedures in the variational approximation corresponding to  mean-field approximation and loopy belief propagation. Essentially, we use neural network to unroll the fixed point update rule of the variational inference (e.g., equation \eqref{equ:fix_point} and \eqref{equ:message_passing}  $m$ steps.
	}
	\label{fig:architecture}
	\vspace{-5mm}
\end{figure*}

\emph{The organization of the following sections}: we prove that the optimal policy has  a Markov Random Field (MRF) form, which reduces the exponential large searching space to a polynomial one. However implement a MRF policy is not trivial in the RL setting.  We resort to the variational inference and derive a fixed-point update rule. At last, we device the variational inference layers to represent the update rule so that the whole system is differentiable (see the architecture in Figure \ref{fig:architecture}).

\vspace{-2mm}
\subsection{Reduce Policy Searching Space}\label{section:reduce_policy_space}
\vspace{-2mm}
Recall that our aim is to maximize the long term reward with the entropy term. Therefore, we follow the definition of the optimal policy in the probabilistic reinforcement learning in \citep{levine2018reinforcement} and obtain the proposition \ref{proposition:MRF}.  It says under the assumption  $r_i(s,\mathbf{a})=r_i(s,a_i, a_{\mathcal{N}_i})$,  the optimal policy in  probabilistic RL is in the form of Markov Random Field~(MRF).
We prove the  proposition in \ref{section:proof}.

\begin{proposition}\label{proposition:MRF}
	Under the assumption  $r_i(s,\mathbf{a})=r_i(s,a_i, a_{\mathcal{N}_i})$, the optimal policy in probabilistic RL has the form $\pi^*(\mathbf{a}^t|s^t)=\frac{1}{Z}\exp(\sum_{i=1}^{N} \psi_i(s^t, a_i^t, a_{\mathcal{N}_i}^t)),$ where $Z$ is the normalization term.
\end{proposition}
\vspace{-2mm}

This proposition is important since it suggests that we should construct the policy  $\pi(\mathbf{a}^t|s^t)$ with this form to contain the optimal policy. If  agent $i$ and its neighbors compose a clique, the policy reduces to a MRF and $\psi_i$ is the potential function. One common example is that the reward is a function on pairwise actions, i.e.,  $r(s,\mathbf{a})=\sum_{i\in \mathcal{V}}r(s,a_i)+\sum_{(i,j) \in \mathcal{E}} r(s,a_i,a_j)$. Then the policy has the form $$\pi(\mathbf{a}|s)=\frac{1}{Z}\exp(\sum_{i\in \mathcal{V}}\psi_i(s,a_i) +\sum_{(i,j)\in \mathcal{E}} \psi_{ij}(s,a_i,a_j)),$$ which is the pairwise MRF. For instance, in traffic lights control, we can naturally define a 2-D grid network and the pairwise reward function.  The MRF formulation on the policy effectively reduces the policy space comparing with the \emph{exponentially} large one in the fully connected graph.

A straightforward way to leverage such observation is to define a
$\pi_\theta(\mathbf{a}^t|s^t)$ as a MRF, and then apply the policy gradient algorithm, e.g., the following way in SAC.
$$\nabla_\theta \mathbb{E}_{s^t\sim D}\mathbb{E}_{a^t\sim \pi_\theta}[\log \pi_\theta(\mathbf{a}^t|s^t) -Q_\kappa(s^t,\mathbf{a}^t) ].$$
However it is still very \emph{hard} to sample joint action $\mathbf{a}^t$ from a MRF  and  two well-known tools are MCMC and variational inference \citep{bishop2006pattern}.  MCMC can generate the unbiased samples from MRF \emph{asymptotically} but it is computationally expensive. As such, we resort to the variational inference.

\subsection{Variational Inference for MRF}

Variational inference is a generic method to approximate complicated  distribution \citep{bishop2006pattern}.  When combined with deep neural networks, it achieves great success such as generating high fidelity images \citep{kingma2019introduction,van2017neural}. As such we apply variational inference on the MRF policy and device differentiable layers to represent optimization procedure in the variational inference.   Our high-level idea can be decomposed into \emph{two steps}: \textbf{1)}. we derive a fixed point update rule for the variational distribution (e.g., mean-field approximation or belief propagation). \textbf{2)} We use neural networks to unroll this update rule ( section \ref{section:neural_embedd_vi}). In the following, we  demonstrate the first step.

In the mean-field variational approximation, we hope to approximate  $\pi(a|s)$ by variational distribution family $p_i$, which is $\min_{(p_1,p_2,...,p_N)} KL (\prod_{i=1}^{N} p_i(a_i|s)|| \pi(\mathbf{a}|s))$ subject to $\int p_i(a_i|s)da_i =1$.  \footnote{It has an equivalent free energy form and we defer it to appendix \ref{app:free_energy_mean}} In the loopy belief propagation, the objective  would be more complicated, where we would use variational distribution $q_i,q_{ij}$ to approximate $\pi(a|s)$ by minimizing the Bethe free energy, i.e., $\min_{\{q_i,q_{ij}\}{(i,j)\in \mathcal{E}}} Bethe(\pi,\{q_i\},\{q_{ij}\})$ with some constraints over the distributions $q_i,q_{ij}$ \citep{yedidia2001bethe}. Remarks that we do not know  the true value of $\psi_i$ and $\psi_{ij}$ in MRF.  We approximated them by neural networks and hope they are learned by the (policy) gradient (The Figure \ref{fig:architecture}). However,  above constrained optimization problem  would \emph{block} the gradient. To circumvent such issue, we present optimal solution of above optimization problem in a fixed-point iteration \cite{blei2016variational}. From a high level, the variational inference layers in Figure \ref{fig:architecture} unroll this fixed-point expression using neural networks so the overall policy is differentiable.  To ease the exposition, we just demonstrate the case in the mean-field approximation and defer more comprehensive discussion on other approximations to Appendix \ref{section:other_variational_inference}.

We denote the optimal solution of the mean-field approximation as $q_i$. Using the coordinate ascent, we obtain the optimal solution $q_i$ should satisfy the following fixed point equation in the mean-field approximation \citep{bishop2006pattern}.

\begin{equation}\label{equ:mean_field_solution}
q_i(a_i|s) \propto \exp \int \prod_{j\neq i} q_j(a_j|s) \log \pi(\mathbf{a}|s)d\mathbf{a}. 
\end{equation}


Recall the policy is a MRF, i.e.,
$$\pi(\mathbf{a}|s)=\frac{1}{Z}\exp(\sum_{i\in\mathcal{V}} \psi_i(s,a_i)+\sum_{(i,j)\in \mathcal{E}} \psi_{ij}(s,a_i,a_j)).$$
Here for simplicity of the representation,  we assume that the MRF is pairwised but the methodology applies to the general case with more involved expression.
We plug this into \eqref{equ:mean_field_solution} and obtain
following fixed point equation.
\begin{flalign}\label{equ:fix_point}
\begin{aligned}
\log q_i(a_i|s)\leftarrow c_i+ \psi_i(s,a_i)
+\sum_{j\in \mathcal{N}_i}\int q_j(a_j|s)\psi_{ij}(s,a_i,a_j)da_j, 
\end{aligned}
\end{flalign}
where $c_i$ is some constant that does not depend on $a_i$.  Remark that so far we have not made any parametric assumption on $q_i$. What we do is to seek functions $q_i$ (optimal of $p_i$) that minimize the functional $KL (\prod_{i=1}^{N} p_i(a_i|s)|| \pi(\mathbf{a}|s))$. Also note that \eqref{equ:fix_point} basically says each agent can \emph{not make the decision independently}. Instead its policy $q_i$ should depend on the policies of others, particularly the neighbors in the equation. 

Clearly, if we unroll \eqref{equ:fix_point} and  design neural network layers corresponding to the term $\psi$, $q$ and summation operation, the final output  would be the mean-field approximation of the joint policy. However we can \emph{not} directly apply this update in our algorithm, since it includes a complicated integral. To this end, in the next section  we resort to the embedding of the distribution $q_i$ \citep{smola2007hilbert}, which says we can represent the distribution by embeddings under mild conditions. Therefore we can absorb the integral  into some operators which will be learned by the neural networks.

\subsection{Neural Embedded Variational Inference}\label{section:neural_embedd_vi}
{\bf Embed the update rule.}
Observe that the fixed point formulation \eqref{equ:fix_point} says that $q_i(a_i|s)$ is a functional of neighborhood marginal distribution $\{q_j(a_j|s)\}_{j\in \mathcal{N}_i}$, which is
\begin{equation}\label{equ:functional_neighbor}
q_i(a_i|s)=f(a_i,s, \{q_j\}_{j\in \mathcal{N}_i}).
\end{equation}
In the following, we denote the d-dimensinoal embedding of  $q_j(a_j|s)$ by $\Tilde{\mu}_j=\int q_j(a_j|s) \phi(a_j|s)da_j$ (section \ref{section:background}).
Notice  the form of feature $\phi$ is not fixed at the moment and will be learned implicitly by the neural network and policy gradient. Following the assumption that there exists a feature space such that the embeddings are injective in Section \ref{section:background}, we can replace the distribution $q_j$ on the right hand side of \eqref{equ:functional_neighbor} by its  embedding $\tilde{\mu}_j$ and have the fixed point formulation as 
\begin{equation}\label{equ:output_distribution}
q_i(a_i|s)=\Tilde{f}(a_i,s,\{\Tilde{\mu}_j\}_{j\in{\mathcal{N}_i}}). 
\end{equation}
Then we multiply the feature map $\phi$ at both sides and do integration over action $a_i$, which yields, $  \Tilde{\mu}_i\leftarrow \int \Tilde{f}(a_i,s,\{\Tilde{\mu}_j\}_{j\in{\mathcal{N}_i}})\phi(a_i|s)da_i$ (recall $\tilde{\mu}_i$ $=\int q_i(a_i|s) \phi(a_i|s)da_i$).  Thus we can rewrite it as a new operator on the  embedding, which induces a fixed point equation again $\Tilde{ \mu }_i \leftarrow \Tilde{\mathcal{T}}\circ (s, \{\Tilde{\mu}_j\}_{j\in \mathcal{N}_i})$.
Roughly speaking, we replace the $q_i$ by $\tilde{\mu}_i$, and then absorb the integral operation and dependence on $\psi_i,\phi$ into $\tilde{\mathcal{T}}$.
Also notice that $\mu_i$ depends on the distribution embedding of its neighbors, which is $\{\tilde{\mu}_j\}_{j_\in \mathcal{N}_i}$.

In practice, we unroll this fix-point expression $M$ iterations.
\begin{equation}\label{equ:message_passing}
\Tilde{\mu}^{m}_{i}\leftarrow \Tilde{\mathcal{T}}\circ (s, \{\Tilde{\mu}^{m-1}_j\}_{j\in \mathcal{N}_i}) \quad m=1,...,M.
\end{equation}
Figure \ref{fig:architecture} (b) represents the message passing of above equation from $m-1$ th iteration  to $m$ th iteration. Finally, we output the distribution $q_i$ with
\begin{equation}\label{equ:output}
q_i(a_i|s)=\Tilde{f}(a_i,s,\{\Tilde{\mu}^{M}_j\}_{j\in{\mathcal{N}_i}}).
\end{equation}
In the loopy belief propagation, we need to introduce two kinds of embedding $\tilde{\nu}_{ij}$ and $\tilde{\mu}_i$ (Figure \ref{fig:architecture} (c)), which corresponds to $q_{ij}$ and $q_i$ in Bethe energy approximation respectively. The fixed-point update rules are $\tilde{\nu}_{ij} \leftarrow\tilde{\mathcal{T}}_1\circ \big( s, \{ \tilde{\nu}_{ki}\}_{k\in \mathcal{N}_i \setminus j}  \big)$,
$ \tilde{\mu}_i \leftarrow \tilde{\mathcal{T}}_2\circ \big( s, \{\tilde{\nu}_{ki}\}_{k\in \mathcal{N}_i} \big)$ (Appendix \ref{section:other_variational_inference}). In the next section, we show how to represent these function and operator by neural networks.

{\bf Parameterization by  Neural Networks.} In general, $\Tilde{\mathcal{T}}$ and $\Tilde{f}$ in  \eqref{equ:message_passing} and \eqref{equ:output} have complicated dependency on $\psi$ and $\phi$. Instead of deriving such dependency, we can directly approximate $\Tilde{f}$ and $\Tilde{\mathcal{T}}$ by neural networks and learn them from data. We can apply  any recent progress in deep learning such as multi-head attention \citep{vaswani2017attention}  on these networks. A \emph{naive} parameterization of $\Tilde{\mathcal{T}}$ in \eqref{equ:message_passing} is an MLP, i.e., $\Tilde{\mu}_i=\sigma(W_1s+W_2\sum_{j\in \mathcal{N}_i}\Tilde{\mu}_j) $, where $\sigma$ is a nonlinear activation function.  This simple case would \emph{reduce} into Graph Neural Network (GNN) \citep{hamilton2017inductive}. Recall $\tilde{f}$  in \eqref{equ:output} is a mapping from the embedding $\mu_i$ to the distributions of action $a_i$. If the distribution is the discrete, $\tilde{f}$ is a softmax function. While the actions space is continuous one, we use the reparameterization trick and output the mean and standard deviation of Gaussian distribution \cite{kingma2013auto}. Therefore, we can obtain each individual action $a_i$ easily.   Remark that this neural embedded variational inference is differentiable.

In the loopy belief propagation,  a simple realization may be $\tilde{\nu}_{ij} =\sigma\big(  W_1s+ W_2 \sum_{k\in \mathcal{N}_i \setminus j} \tilde{\nu}_{ki} \big),$ $ \tilde{\mu}_i= \sigma \big( W_3s+W_4 \sum_{k\in \mathcal{N}_i} \tilde{\nu}_{ki} \big).$  To the best of our knowledge, such neural belief propagation message passing form is new in MARL community. Again, we can approximate $\tilde{\mathcal{T}}_1, \tilde{\mathcal{T}}_2$  by any other complicated networks. We demonstrate the details of our parameterization  in \textbf{Appendix} \ref{app:detail_intention_propgation_network}. 

\subsection{Algorithm}\label{section:algorithm}


We give the overall architecture  by combining all pieces together in Figure \ref{fig:architecture}. In the forward pass, the agent (policy) observes the state information. We use the neural networks to approximate $\psi_i$ and $\psi_{ij}$  whose embeddings then pass through the variational inference layers and obtain the variational approximation $q_i$ in \eqref{equ:output}. We then sample $a_i$ from $q_i$. We can define the $V_i(s)$ and $Q_i(s,a_i,a_{\mathcal{N}_i})$ for each agent i according to this variational approximation, where each of them is approximated by the neural network with parameter $\eta_i$ and $\kappa_i$ respectively. We have a target network of $V_i$ with parameter $\Bar{\eta}_i$ as the common practice in off-policy reinforcement learning \cite{haarnoja2018softac}. We use $\theta$ to represent  the parameters in policy $\pi$, i.e., all parameters of neural network in Figure \ref{fig:architecture}.  Here we  represent the loss function of them  and defer the derivations of them to \ref{section:derivation}.

\begin{equation}\label{equ:loss}
\begin{split}
&J_\pi(\theta)= \mathbb{E}_{s^t\sim D, a_i^t\sim q_i} \sum_{i=1}^{N}[\log q_{i,\theta}(a^t_i|s^t)- Q_{\kappa_i}(s^t, a_i^t, a_{\mathcal{N}_i}^t)]\\
&J_V(\eta_i)=\mathbb{E}_{s^t\sim D}[\frac{1}{2}\big(V_{\eta_i}(s^t)-
\mathbb{E}_{(a_i^t,a^t_{\mathcal{N}_i)}\sim (q_i,q _{\mathcal{N}_i})}  [Q_{\kappa_i}(s^t,a_i^t,a^t_{\mathcal{N}_i} )-\log q_{i,\theta}(a_i^t|s^t)]  \big)^2 ].\\  &J_Q(\kappa_i)=\mathbb{E}_{(s^t,a_i^t,a^t_{\mathcal{N}_i})\sim D} [\frac{1}{2}\big( Q_{\kappa_i}(s^t,a_i^t, a_{\mathcal{N}_i}^t)-\hat{Q}_i(s^t, a_i^t,a_{\mathcal{N}_i}^t) \big)^2],\\
&\hat{Q}_i(s^t,a_i^t, a^t_{\mathcal{N}_i})=r_i+\gamma \mathbb{E}_{s^{t+1}\sim p(\cdot|s_t,\mathbf{a}^t)}[V_{\Bar{\eta}_i}(s^{t+1})].\\
\end{split}
\end{equation}
In above equations $D$ stands for the replay buffer. The procedure of the whole algorithm runs as follows. The agent samples the action from the MRF  and impose it on the environment. The agent store the tuple $(s_i^t,a_i^t,r_i^t,s_i^{t+1})_{i=1}^N$ to the replay buffer $D$. In every step, we update the parameter of $\pi$,$V_i$ $Q_i$ using \eqref{equ:loss}. The pseudo-code of  algorithm in represented in appendix \ref{section:algorithm}.

It is interesting to compare the loss  with the counterpart in the single agent SAC in section \ref{section:background}. 
\begin{itemize}[leftmargin=*,nolistsep,nosep]
	\item $q_{i,\theta}(a_i|s)$ is the output of variational inference layers. It depends on the policy of other agents.
	\item $Q_{\kappa_i}$ depends on the action of itself and its neighbors, which can be accomplished by the graph neural network in practice. \\
\end{itemize}
\vspace{-2mm}

\textbf{Discuss related message-passing work in MARL:}
The coordination graph approach  \citep{bohmer2019deep} is a rigorous work to mitigate the problem of the exponentially large joint action space. It has similar spirit as ours but needs to explicitly calculate the max-sum operator in the message-passing (e.g., belief propagation), which can not fit the large scale and continuous action setting. From a high-level, our work uses neural networks to implicitly approximate such step in the belief propagation and mitigates the scalability issue while maintains the theoretical guarantee. \citet{jiang2020graph} use GNN to spread the state information to the neighbors and learn the individual $Q_i$, which mitigates the partially observable problem. However agents still make independent decision since each agent does not care what others will do. Check that they do not have max-sum or product-sum step on action which  essentially is the core in the coordination graph approach \citep{bohmer2019deep}.  \citet{chu2020multi}  use LSTM to extract state action information and spread it to neighbors. It needs strong assumption that MDP has spatial-temporal Markov property. However, we do not know which point will this message-passing algorithm converge to.  In the experiment, we will see that VPP outperforms the baselines with message-passing. It is because that we build the joint-policy approximation problem as a rigorous optimization problem in variational inference and  construct better message-passing.

\textbf{Handle the Partial Observation:}
So far, we assume that agents can observe global state while in practice, each agent just observes its own state $s_i$. Thus, in VPP (e.g., Figure \ref{fig:network_parameterization} in appendix ), we also do the message passing on state and use $ s_i^{m}$ to replace the global state $s$. We do similar things on the input state of $Q_i$ and $V_i$. There are many other ways to alleviate the partial observation problem, but it is not the focus of this paper.

\section{Experiment}
\label{sec:exp}
\begin{figure}
	\vspace{-3mm}
	\centering
	\includegraphics[width=0.5\textwidth]{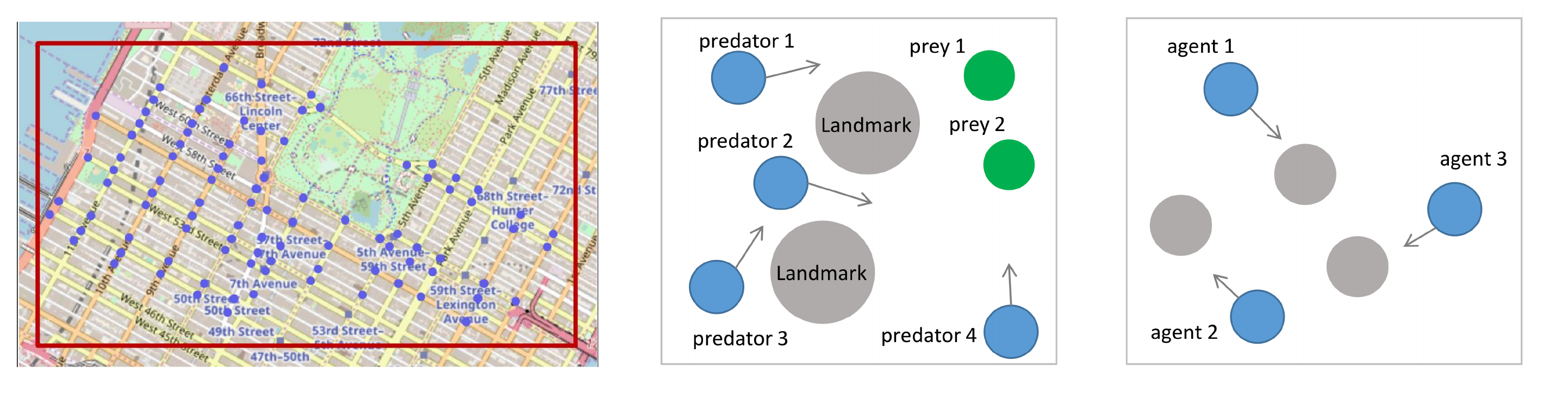}
	\vspace{-3mm}
	\caption{Experimental scenarios. Cityflow: Manhattan, Predator-Prey and Cooperative-Navigation.
	}\vspace{-3mm}
	\label{fig:scenario_figure}
\end{figure}

\begin{figure*}
	\centering
	\subfloat[CityFlow:Manhattan]{\includegraphics[width=.315\textwidth]{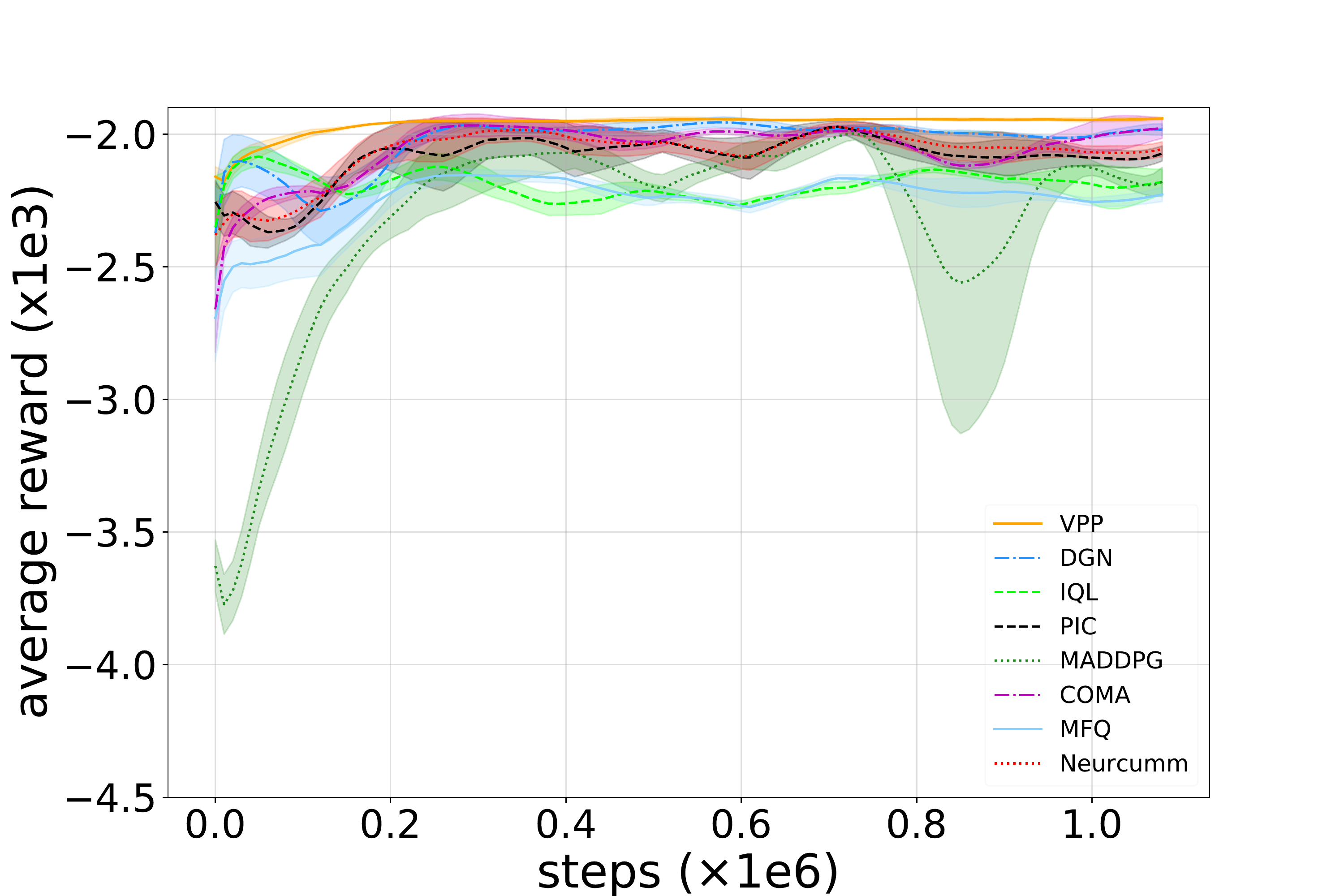}}
	\subfloat[CityFlow:N=100]{\includegraphics[width=.315\textwidth]{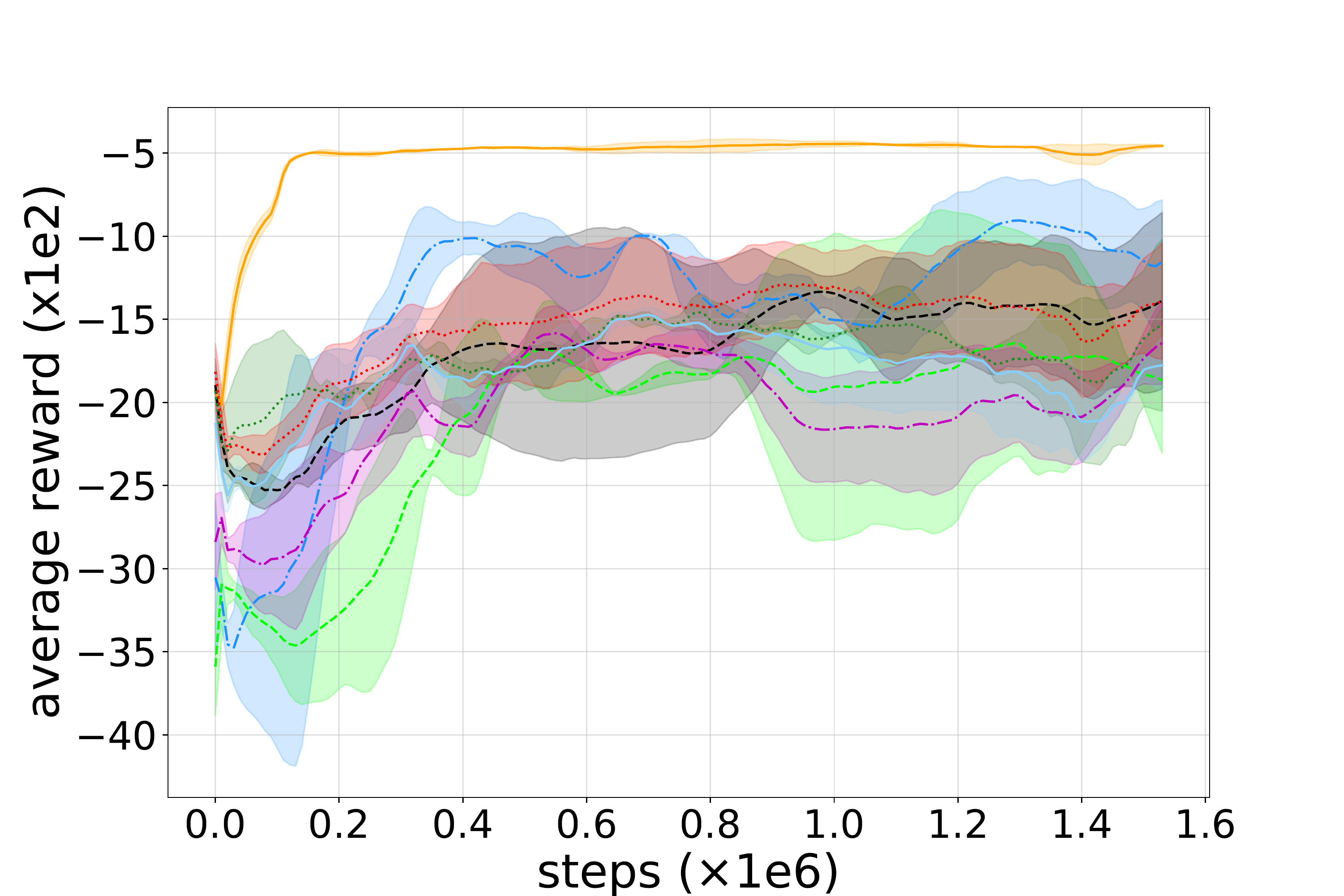}}
	\subfloat[CityFlow:N=1225]{\includegraphics[width=.315\textwidth]{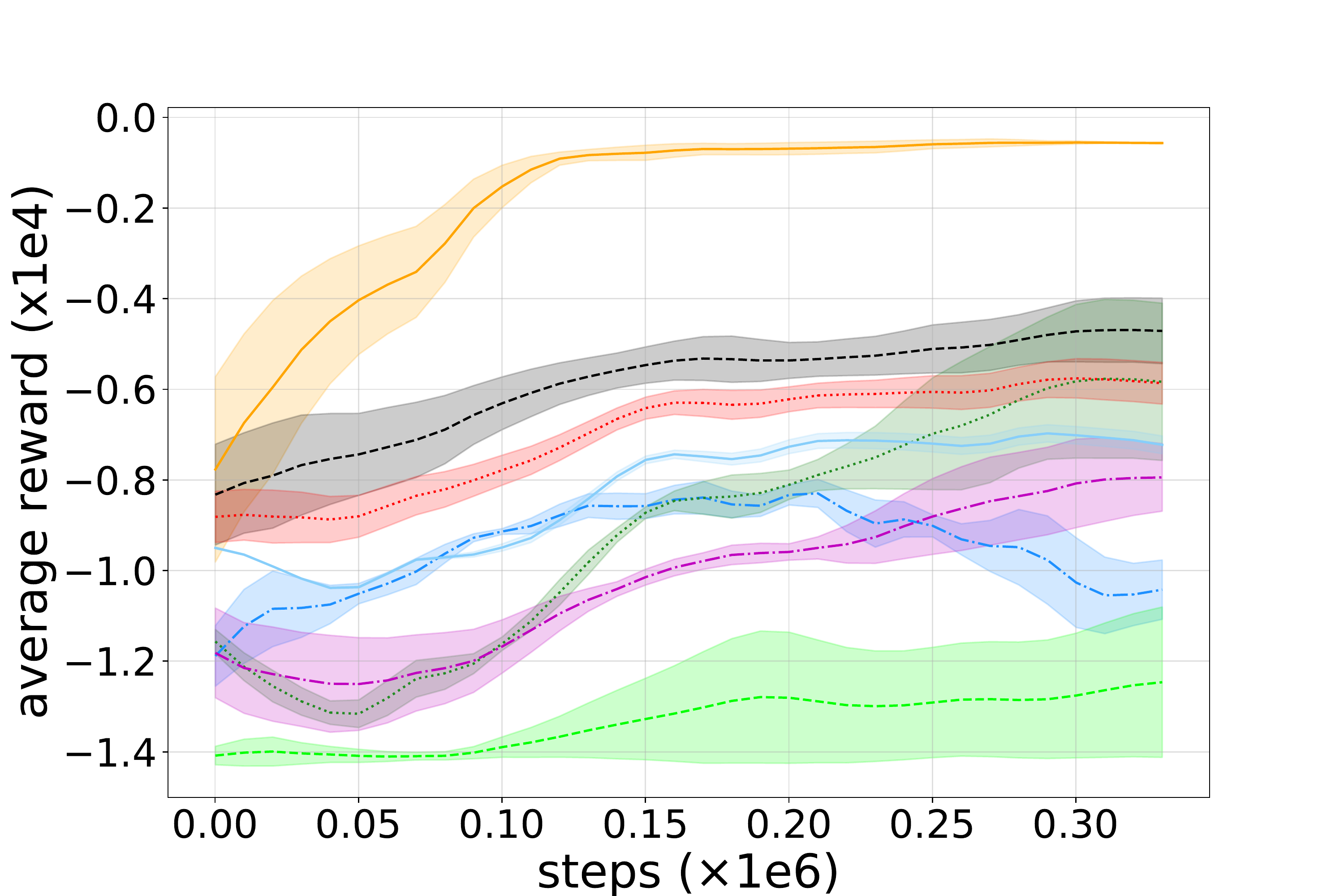}}
	\caption{Performance on large-scale traffic lights control scenarios in CityFlow. 
		Horizontal axis: environmental steps. Vertical axis: average episode reward (negative average travel time).
		Higher rewards are better.
		Our VPP performs best especially on large-scale tasks.
	}\label{fig:exp_cityflow_size}
\end{figure*}

In this section, we evaluate our method and seven state-of-the-art baselines on more than ten different scenarios
from three popular MARL platforms: (1) CityFlow, a traffic signal control environment \citep{tang2019cityflow}. It is an advanced version of SUMO \citep{lopez2018microscopic}  widely used in MARL community. (2) multiple particle environment~(MPE)~\citep{mordatch2017emergence} and (3) grid-world platform MAgent~\citep{zheng2018magent}. We illustrate the experimental scenarios in Figure \ref{fig:scenario_figure}.
In all experiment, we implement our mean-field approximation version of VPP. Our algorithm empirically outperforms all baselines on all scenarios especially on the \emph{large scale} problem. 


\vspace{-2mm}
\subsection{Settings}
\vspace{-2mm}

We give a brief introduction to the settings of the experiment and defer the details such as hyperparameter tuning of VPP and baselines to appendix~\ref{app:env}. Notice all algorithms are tested in the partially observable setting, i.e., each agent just can observe its own state $s_i$.

In traffic signal control problem (left panel in Figure~\ref{fig:scenario_figure}), each traffic light at the intersection is an agent. The goal is to learn policies of traffic lights to reduce the average waiting time to alleviate the traffic jam. \emph{Graph for cityflow}: graph is a 2-D grid induced by the map in Figure \ref{fig:scenario_figure}. The  roads are the edges which connects the agents.  We can define the cost $-r_i$ as the traveling time of vehicle around the intersection $i$, thus the total cost indicates the average traveling time. Obviously, $r_i$ has a close relationship with the action of neighbors of agent $i$ but has little dependence on the traffic lights far away. Therefore our assumption on reward function holds. 
We evaluate different methods on both real-world and synthetic traffic data under the different numbers of intersections.

MPE~\citep{mordatch2017emergence} and MAgent~\citep{zheng2018magent} (right panel in Figure \ref{fig:scenario_figure}) are popular particle environments on MARL~\citep{lowe2017multi, jiang2020graph}.  \emph{Graph for particle environments } : for each agent, it has connections (i.e., the edge of the graph) with  $k$ nearest neighbors. Since the graph is dynamic, we update the adjacency matrix of the graph every $n$ step, e.g., $n=5$ steps. It  just introduce a small overhead comparing with the training of the neural networks.   The reward functions also have local property, since they are explicitly or implicitly affected by the distance between agents. For instance, in heterogeneous navigation,  if small agents collide with big agents, they will obtain a large negative reward. Thus their reward depends on the action of the nearby agents. Similarly, in the jungle environment, agent can attack the agents nearby  to obtain a high reward.

\textbf{Baselines.} 
We compare our method against seven different baselines mentioned in the introduction and related work sections:
MADDPG~\citep{lowe2017multi}; 
permutation invariant critic (PIC)~\citep{liu2019pic};
graph convolutional reinforcement learning~(DGN)~\citep{jiang2020graph};
Neurcomm \citep{chu2020multi}; COMA \citep{foerster2018counterfactual}; MFQ \citep{yang2018mean}; Independent Q learning (IQL) \citep{tan1993multi};
These baselines are reported as the leading algorithm of solving tasks in CityFlow, MPE and MAgent. Among them, DGN, Neurcomm and MFQ need the communication with neighbors in the training and execution as ours. MFQ is an algorithm considering the mean action approximation and thus it is interesting to compare this approximation method with our variational inference approximation. Also notice that PIC assumes the actor can observe the global state. Thus in the partially observable setting, each agent in PIC also needs to communicate to get the global state information in the training and the execution. There are also other works related to our VPP such  as \citep{bohmer2019deep,wen2018probabilistic}. However they can not fit the large-scale setting (e.g., $n>30$) due to the computational issue. Thus we do not report them here.  Further details on baselines are given in appendix~\ref{app:baseline}.

\textbf{Neural Network and Parameters.} 
Recall that in the variational inference layers (section \ref{section:neural_embedd_vi}), we can approximate the operator $\tilde{\mathcal{T}}$ by neural networks. In particular, for each agent we stack two  multi-head attention+MLP modules  (over itself and its neighbors)  to represent two rounds of  mean-field message-passing. Then the output feeds into a MLP+softmax layer. The stucture and hyperparameters are deferred in appendix~\ref{app:detail_intention_propgation_network},\ref{app:hypers}.

\begin{figure*}
	\vspace{-5mm}
	\centering
	\subfloat[Jungle(N=20, Food=12)]{\includegraphics[width=.315\textwidth]{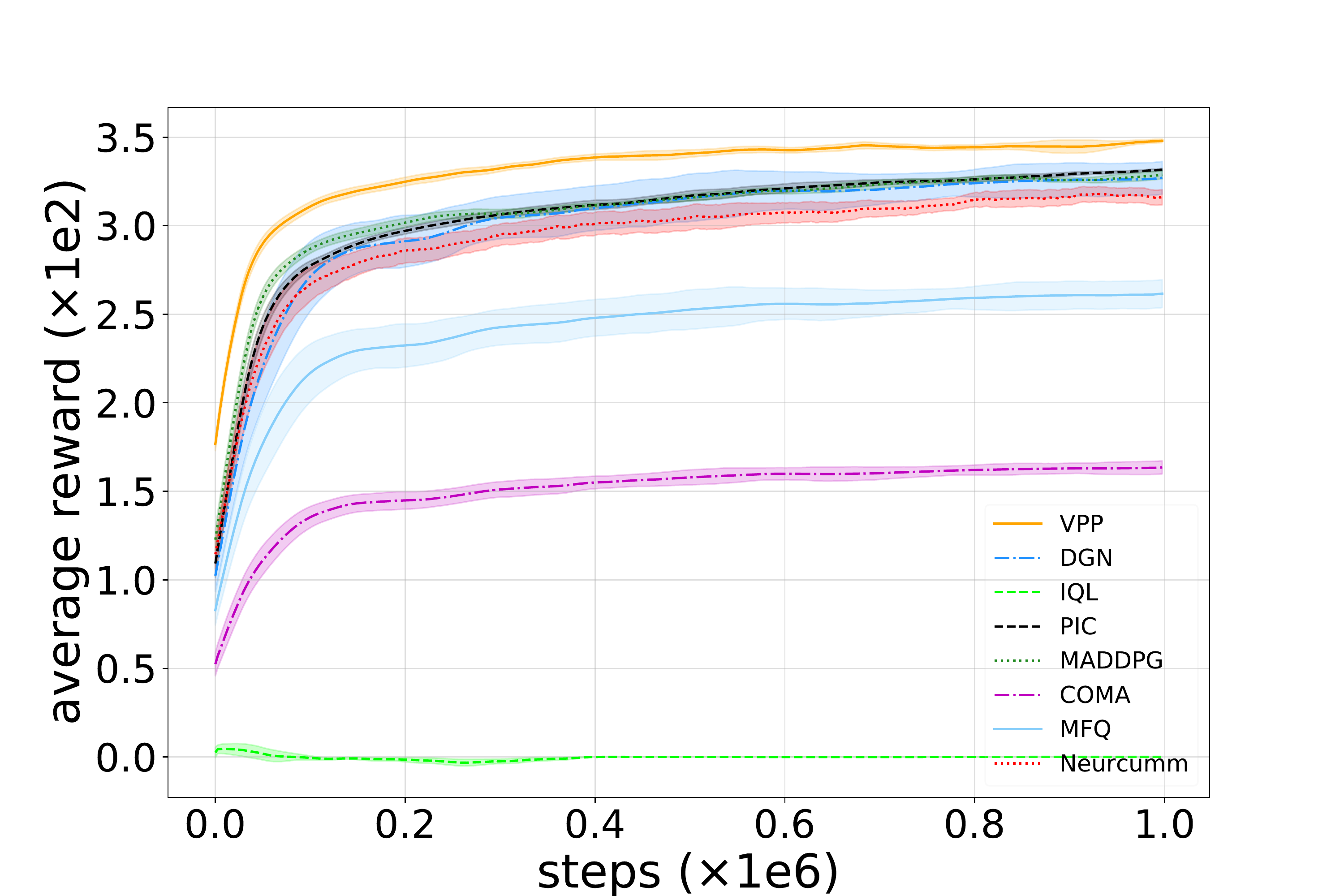}}\quad
	\subfloat[Cooperative Nav (N=30)]{\includegraphics[width=.315\textwidth]{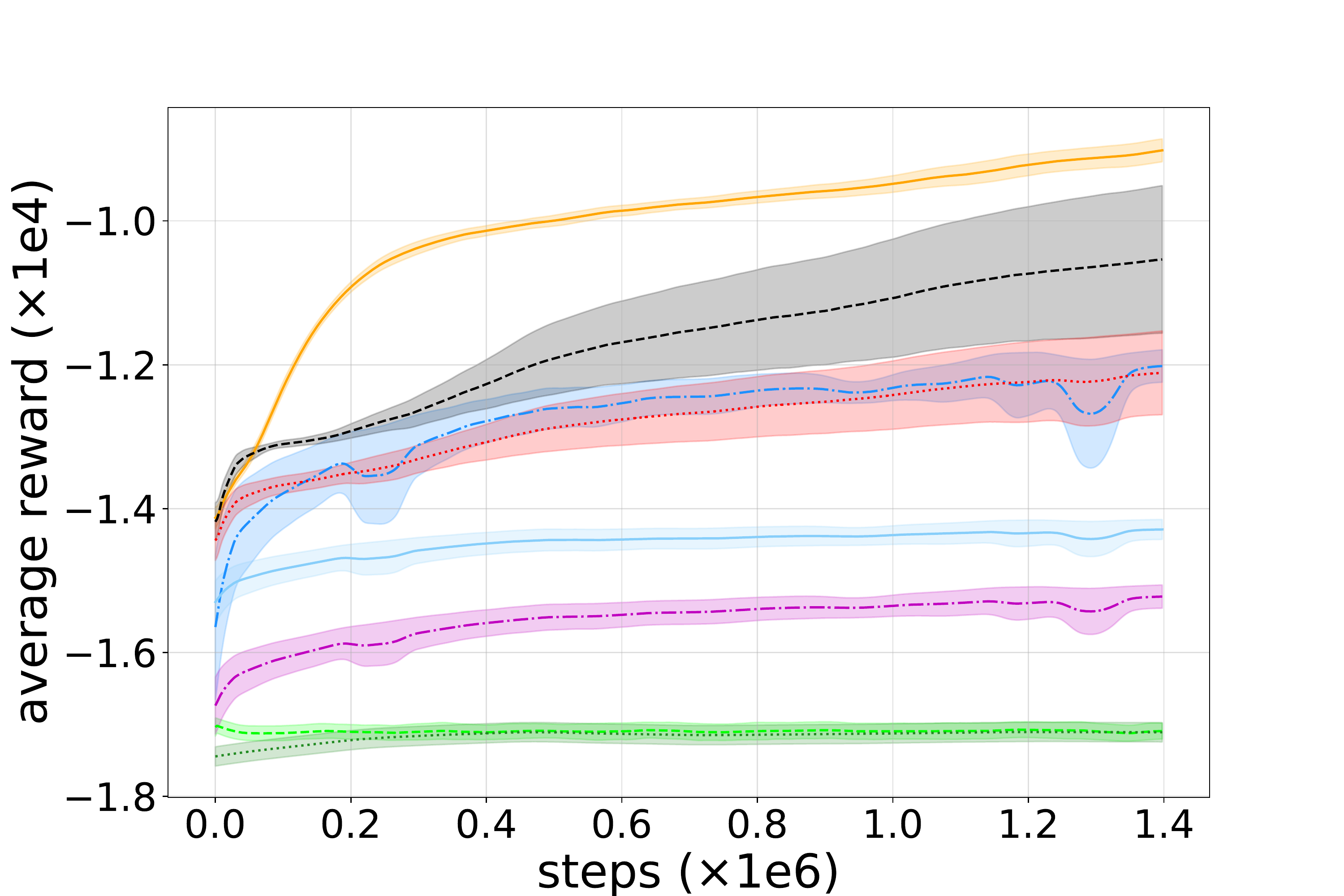}}\quad
	\subfloat[ Heterogeneous Nav (N=100)]{\includegraphics[width=.315\textwidth]{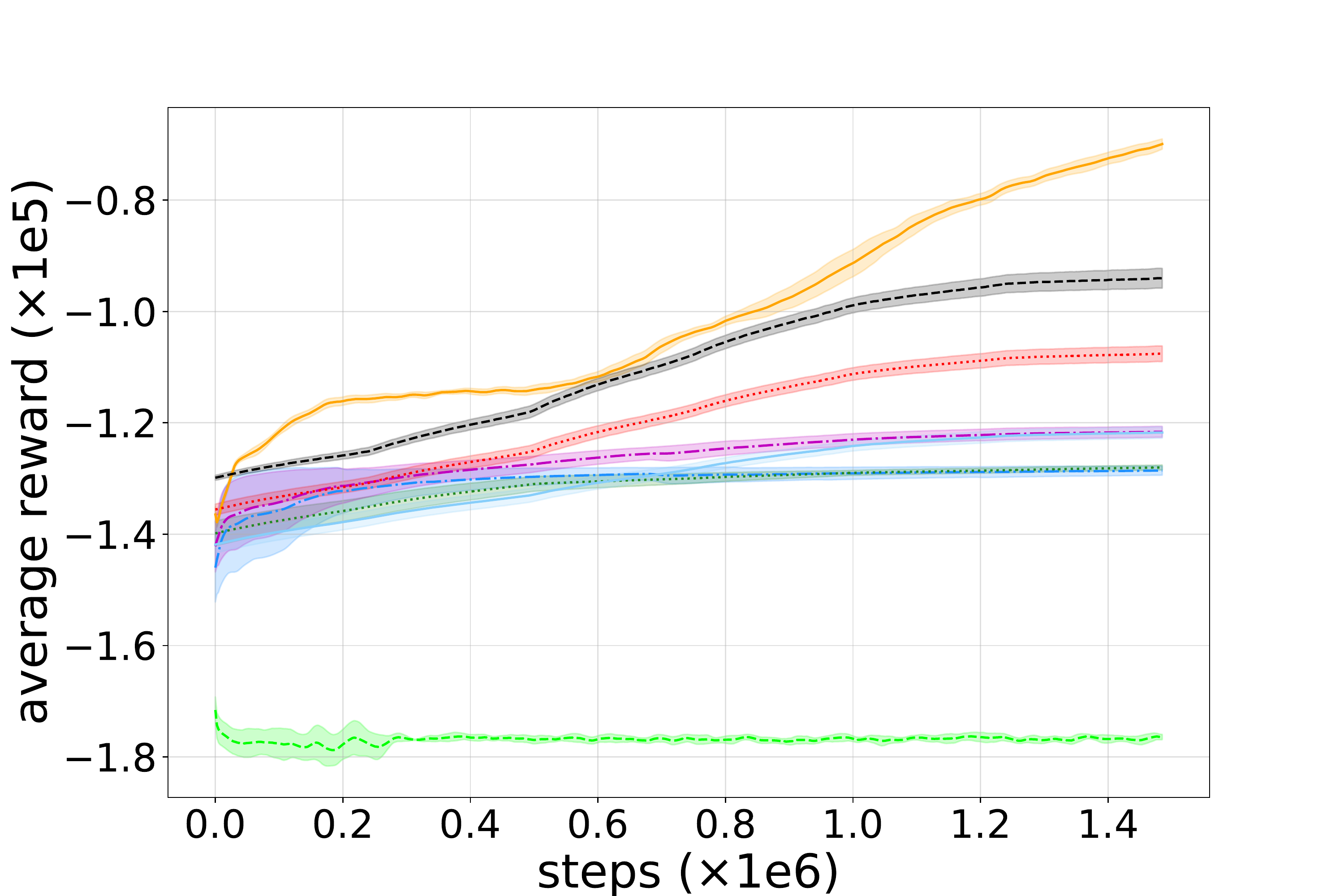}}\quad
	\vspace{-2mm}
	\caption{Experimental results on Jungle, Cooperative Navigation and Heterogeneous Navigation. VPP beats all the baselines. 
	} \vspace{-3mm}
	\label{fig:exp_2}
\end{figure*}


\vspace{-2mm}
\subsection{Comparison to state-of-the-art}

In this section, we compare VPP with other baselines.  The experiments are evaluated by average episode reward~\citep{lowe2017multi}.
For CityFlow tasks, average reward refers to negative average travel time.
All experiments are repeated for 5 runs with different random seeds. We report the mean and standard deviation in the curves. We report the results on six experiments and defer all the others to \textbf{appendix} \ref{app:further_exps} due to the limit of space.  

\textbf{CityFlow.}
We first evaluate our algorithm on traffic control problem. Specifically, we increase the number of  intersections (agents) gradually to increase the difficulties of the tasks. Figure~\ref{fig:exp_cityflow_size} presents the performance of different methods on both real-world and synthetic CityFlow data with different number of intersections. On the task of Manhattan City,  VPP and the baseline methods such as PIC, DGN and Neurcomm achieve better reward than the other methods while our method approaches higher reward within fewer steps.
On the larger task (N=100), VPP outperforms PIC and DGN a lot. Further more the baselines have larger variance.
The experiment with N=1225 agents is an extremely challenging task. Our algorithm outperforms all baselines with a wide margin. The runner-up is PIC whose final performance is around  $-4660$ and suffers from large variance. In contrast, the performance of our method is around $-569$ (much higher than the baselines). It's clear that, in both real-world and synthetic cityflow scenarios, the proposed VPP method obtains the best performance.

\textbf{MPE and MAgent.}
Figure~\ref{fig:exp_2} demonstrates the performance of different methods on other three representative scenario instances:   \textit{jungle} (N=20,F=12), \textit{cooperative navigation} (N=30) and \textit{heterogeneous navigation} (N=100). We run all algorithms long enough (more than 1e6 steps). In all experiments, our algorithm performs best.
In \textit{cooperative navigation N=30}, the runner up is PIC and it is followed by DGN and Neurcumm. When solving large-scale settings (heterogeneous navigation N=100), the performance of DGN and Neurcumm is much worse than PIC and our VPP.
Although PIC can solve large-scale tasks, VPP is still much better. 


\textbf{Stability.} Stability is a key criterion to evaluate MARL. In all experiments, our method is quite stable with small variance.  As shown in Figure \ref{fig:exp_cityflow_size} (b), DGN approaches $-1210 \pm 419$ on the CityFlow scenario with N=100 intersections while our method approaches $-465 \pm 20$ after $1.6\times 10^6$ steps (much better and stable). The reason is that VPP can make the better approximation on the joint policy through the variational inference while other methods may converges to the policy which is far from the joint policy even they also consider the messages from the neighbors. 

\subsection{Ablation Study}
We conduct a set of ablation studies related to the effect of joint policy, graph, rounds of the message-passing, number of neighbors, time interval to update adjacency matrix and the assumption of the reward function.  In particular, we find the joint policy is essential for the good performance. In Cityflow, the performance of  traffic graph (2-d grid induced by the roadmap) is better than the fully connected graph. In MPE and MAgent, We define the adjacent matrix based on the $k$ nearest neighbors and pick $k=8$ in large scale problem and $k=4$ in small scale problem. In all of our experiment, we choose the 2 round message-passing and find that it is enough to approximate the joint policy in most of scenarios.  To save the computation updating the adjacency matrix, we update the matrix every 5 rounds.   Because of the limitation of space, we just summarize our conclusion above and place the details in appendix~\ref{app:ablation}.

\section{Conclusion}
In this paper, we propose a framework to approximate the joint policy efficiently in the collaborative MARL problem inspired by the variational inference . In theory, such neural message-passing algorithms converge approximately to the fixed point of the mean-field, loopy belief propagation and many other famous variational inference methods. In practice, it outperforms several baselines especially in the large-scale problem. One limitation of our work is that we use the continuous embeddings as the messages. In future work, we may discretize or add a decoder to map the continuous messages to discrete symbols to mitigate bandwidth problem. Another direction is to improve the communication efficiency, e.g., design or learn the communication protocol such that the agent communicate when it is necessary.

\bibliography{MARL}

\begin{thebibliography}{51}
\providecommand{\natexlab}[1]{#1}
\providecommand{\url}[1]{\texttt{#1}}
\expandafter\ifx\csname urlstyle\endcsname\relax
  \providecommand{\doi}[1]{doi: #1}\else
  \providecommand{\doi}{doi: \begingroup \urlstyle{rm}\Url}\fi

\bibitem[Amos and Kolter(2017)]{amos2017optnet}
Brandon Amos and J~Zico Kolter.
\newblock Optnet: Differentiable optimization as a layer in neural networks.
\newblock In \emph{International Conference on Machine Learning}, pages
  136--145. PMLR, 2017.

\bibitem[Bishop(2006)]{bishop2006pattern}
Christopher~M Bishop.
\newblock \emph{Pattern recognition and machine learning}.
\newblock springer, 2006.

\bibitem[Blei et~al.(2016)Blei, Ranganath, and Mohamed]{blei2016variational}
David Blei, Rajesh Ranganath, and Shakir Mohamed.
\newblock Variational inference: Foundations and modern methods.
\newblock \emph{NIPS Tutorial}, 2016.

\bibitem[B{\"o}hmer et~al.(2020)B{\"o}hmer, Kurin, and
  Whiteson]{bohmer2019deep}
Wendelin B{\"o}hmer, Vitaly Kurin, and Shimon Whiteson.
\newblock Deep coordination graphs.
\newblock \emph{ICML 2020}, 2020.

\bibitem[Castellini et~al.(2019)Castellini, Oliehoek, Savani, and
  Whiteson]{castellini2019representational}
Jacopo Castellini, Frans~A Oliehoek, Rahul Savani, and Shimon Whiteson.
\newblock The representational capacity of action-value networks for
  multi-agent reinforcement learning.
\newblock In \emph{Proceedings of the 18th International Conference on
  Autonomous Agents and MultiAgent Systems}, pages 1862--1864. International
  Foundation for Autonomous Agents and Multiagent Systems, 2019.

\bibitem[Chu et~al.(2020)Chu, Chinchali, and Katti]{chu2020multi}
Tianshu Chu, Sandeep Chinchali, and Sachin Katti.
\newblock Multi-agent reinforcement learning for networked system control.
\newblock \emph{arXiv preprint arXiv:2004.01339}, 2020.

\bibitem[Dai et~al.(2016)Dai, Dai, and Song]{dai2016discriminative}
Hanjun Dai, Bo~Dai, and Le~Song.
\newblock Discriminative embeddings of latent variable models for structured
  data.
\newblock In \emph{International conference on machine learning}, pages
  2702--2711, 2016.

\bibitem[de~Witt et~al.(2019)de~Witt, Foerster, Farquhar, Torr, Boehmer, and
  Whiteson]{de2019multi}
Christian~Schroeder de~Witt, Jakob Foerster, Gregory Farquhar, Philip Torr,
  Wendelin Boehmer, and Shimon Whiteson.
\newblock Multi-agent common knowledge reinforcement learning.
\newblock In \emph{Advances in Neural Information Processing Systems}, pages
  9924--9935, 2019.

\bibitem[Ding et~al.(2020)Ding, Huang, and Lu]{ding2020learning}
Ziluo Ding, Tiejun Huang, and Zongqing Lu.
\newblock Learning individually inferred communication for multi-agent
  cooperation.
\newblock \emph{Neurips 2020}, 2020.

\bibitem[Donti et~al.(2017)Donti, Amos, and Kolter]{donti2017task}
Priya~L Donti, Brandon Amos, and J~Zico Kolter.
\newblock Task-based end-to-end model learning in stochastic optimization.
\newblock \emph{NIPS2017}, 2017.

\bibitem[Donti et~al.(2021)Donti, Rolnick, and Kolter]{donti2021dc3}
Priya~L Donti, David Rolnick, and J~Zico Kolter.
\newblock Dc3: A learning method for optimization with hard constraints.
\newblock \emph{ICLR2021}, 2021.

\bibitem[Foerster et~al.(2016)Foerster, Assael, de~Freitas, and
  Whiteson]{foerster2016learning}
Jakob Foerster, Ioannis~Alexandros Assael, Nando de~Freitas, and Shimon
  Whiteson.
\newblock Learning to communicate with deep multi-agent reinforcement learning.
\newblock In \emph{Advances in Neural Information Processing Systems}, pages
  2137--2145, 2016.

\bibitem[Foerster et~al.(2018)Foerster, Farquhar, Afouras, Nardelli, and
  Whiteson]{foerster2018counterfactual}
Jakob~N Foerster, Gregory Farquhar, Triantafyllos Afouras, Nantas Nardelli, and
  Shimon Whiteson.
\newblock Counterfactual multi-agent policy gradients.
\newblock In \emph{Thirty-Second AAAI Conference on Artificial Intelligence},
  2018.

\bibitem[Guestrin et~al.(2002)Guestrin, Lagoudakis, and
  Parr]{guestrin2002coordinated}
Carlos Guestrin, Michail Lagoudakis, and Ronald Parr.
\newblock Coordinated reinforcement learning.
\newblock In \emph{ICML}, volume~2, pages 227--234. Citeseer, 2002.

\bibitem[Haarnoja et~al.(2018{\natexlab{a}})Haarnoja, Zhou, Abbeel, and
  Levine]{haarnoja2018soft}
Tuomas Haarnoja, Aurick Zhou, Pieter Abbeel, and Sergey Levine.
\newblock Soft actor-critic: Off-policy maximum entropy deep reinforcement
  learning with a stochastic actor.
\newblock \emph{arXiv preprint arXiv:1801.01290}, 2018{\natexlab{a}}.

\bibitem[Haarnoja et~al.(2018{\natexlab{b}})Haarnoja, Zhou, Hartikainen,
  Tucker, Ha, Tan, Kumar, Zhu, Gupta, Abbeel, et~al.]{haarnoja2018softac}
Tuomas Haarnoja, Aurick Zhou, Kristian Hartikainen, George Tucker, Sehoon Ha,
  Jie Tan, Vikash Kumar, Henry Zhu, Abhishek Gupta, Pieter Abbeel, et~al.
\newblock Soft actor-critic algorithms and applications.
\newblock \emph{ICML 2018}, 2018{\natexlab{b}}.

\bibitem[Hamilton et~al.(2017)Hamilton, Ying, and
  Leskovec]{hamilton2017inductive}
Will Hamilton, Zhitao Ying, and Jure Leskovec.
\newblock Inductive representation learning on large graphs.
\newblock In \emph{Advances in neural information processing systems}, pages
  1024--1034, 2017.

\bibitem[Huang et~al.(2006)Huang, Malham{\'e}, Caines, et~al.]{huang2006large}
Minyi Huang, Roland~P Malham{\'e}, Peter~E Caines, et~al.
\newblock Large population stochastic dynamic games: closed-loop mckean-vlasov
  systems and the nash certainty equivalence principle.
\newblock \emph{Communications in Information \& Systems}, 6\penalty0
  (3):\penalty0 221--252, 2006.

\bibitem[Jiang et~al.(2020)Jiang, Dun, and Lu]{jiang2020graph}
Jiechuan Jiang, Chen Dun, and Zongqing Lu.
\newblock Graph convolutional reinforcement learning.
\newblock \emph{ICLR}, 2020.

\bibitem[Kingma and Welling(2013)]{kingma2013auto}
Diederik~P Kingma and Max Welling.
\newblock Auto-encoding variational bayes.
\newblock \emph{arXiv preprint arXiv:1312.6114}, 2013.

\bibitem[Kingma and Welling(2019)]{kingma2019introduction}
Diederik~P Kingma and Max Welling.
\newblock An introduction to variational autoencoders.
\newblock \emph{arXiv preprint arXiv:1906.02691}, 2019.

\bibitem[Kuyer et~al.(2008)Kuyer, Whiteson, Bakker, and
  Vlassis]{kuyer2008multiagent}
Lior Kuyer, Shimon Whiteson, Bram Bakker, and Nikos Vlassis.
\newblock Multiagent reinforcement learning for urban traffic control using
  coordination graphs.
\newblock In \emph{Joint European Conference on Machine Learning and Knowledge
  Discovery in Databases}, pages 656--671. Springer, 2008.

\bibitem[Levine(2018)]{levine2018reinforcement}
Sergey Levine.
\newblock Reinforcement learning and control as probabilistic inference:
  Tutorial and review.
\newblock \emph{arXiv preprint arXiv:1805.00909}, 2018.

\bibitem[Liu et~al.(2019)Liu, Yeh, and Schwing]{liu2019pic}
Iou-Jen Liu, Raymond~A Yeh, and Alexander~G Schwing.
\newblock Pic: Permutation invariant critic for multi-agent deep reinforcement
  learning.
\newblock \emph{Conference on Robot Learning}, 2019.

\bibitem[Lopez et~al.(2018)Lopez, Behrisch, Bieker-Walz, Erdmann,
  Fl{\"o}tter{\"o}d, Hilbrich, L{\"u}cken, Rummel, Wagner, and
  WieBner]{lopez2018microscopic}
Pablo~Alvarez Lopez, Michael Behrisch, Laura Bieker-Walz, Jakob Erdmann,
  Yun-Pang Fl{\"o}tter{\"o}d, Robert Hilbrich, Leonhard L{\"u}cken, Johannes
  Rummel, Peter Wagner, and Evamarie WieBner.
\newblock Microscopic traffic simulation using sumo.
\newblock In \emph{2018 21st International Conference on Intelligent
  Transportation Systems (ITSC)}, pages 2575--2582. IEEE, 2018.

\bibitem[Lowe et~al.(2017)Lowe, Wu, Tamar, Harb, Abbeel, and
  Mordatch]{lowe2017multi}
Ryan Lowe, Yi~Wu, Aviv Tamar, Jean Harb, OpenAI~Pieter Abbeel, and Igor
  Mordatch.
\newblock Multi-agent actor-critic for mixed cooperative-competitive
  environments.
\newblock In \emph{Advances in Neural Information Processing Systems}, pages
  6379--6390, 2017.

\bibitem[Mordatch and Abbeel(2017)]{mordatch2017emergence}
Igor Mordatch and Pieter Abbeel.
\newblock Emergence of grounded compositional language in multi-agent
  populations.
\newblock \emph{arXiv preprint arXiv:1703.04908}, 2017.

\bibitem[Oliehoek et~al.(2016)Oliehoek, Amato, et~al.]{oliehoek2016concise}
Frans~A Oliehoek, Christopher Amato, et~al.
\newblock \emph{A concise introduction to decentralized POMDPs}, volume~1.
\newblock Springer, 2016.

\bibitem[Palmer et~al.(2018)Palmer, Tuyls, Bloembergen, and
  Savani]{palmer2018lenient}
Gregory Palmer, Karl Tuyls, Daan Bloembergen, and Rahul Savani.
\newblock Lenient multi-agent deep reinforcement learning.
\newblock In \emph{Proceedings of the 17th International Conference on
  Autonomous Agents and MultiAgent Systems}, pages 443--451. International
  Foundation for Autonomous Agents and Multiagent Systems, 2018.

\bibitem[Qu et~al.(2019)Qu, Mannor, Xu, Qi, Song, and Xiong]{qu2019value}
Chao Qu, Shie Mannor, Huan Xu, Yuan Qi, Le~Song, and Junwu Xiong.
\newblock Value propagation for decentralized networked deep multi-agent
  reinforcement learning.
\newblock \emph{Neurips 2019}, 2019.

\bibitem[Rashid et~al.(2018)Rashid, Samvelyan, De~Witt, Farquhar, Foerster, and
  Whiteson]{rashid2018qmix}
Tabish Rashid, Mikayel Samvelyan, Christian~Schroeder De~Witt, Gregory
  Farquhar, Jakob Foerster, and Shimon Whiteson.
\newblock Qmix: monotonic value function factorisation for deep multi-agent
  reinforcement learning.
\newblock \emph{ICML 2018}, 2018.

\bibitem[Shalev-Shwartz et~al.(2016)Shalev-Shwartz, Shammah, and
  Shashua]{shalev2016safe}
Shai Shalev-Shwartz, Shaked Shammah, and Amnon Shashua.
\newblock Safe, multi-agent, reinforcement learning for autonomous driving.
\newblock \emph{arXiv preprint arXiv:1610.03295}, 2016.

\bibitem[Smola et~al.(2007)Smola, Gretton, Song, and
  Sch{\"o}lkopf]{smola2007hilbert}
Alex Smola, Arthur Gretton, Le~Song, and Bernhard Sch{\"o}lkopf.
\newblock A hilbert space embedding for distributions.
\newblock In \emph{International Conference on Algorithmic Learning Theory},
  pages 13--31. Springer, 2007.

\bibitem[Son et~al.(2019)Son, Kim, Kang, Hostallero, and Yi]{son2019qtran}
Kyunghwan Son, Daewoo Kim, Wan~Ju Kang, David~Earl Hostallero, and Yung Yi.
\newblock Qtran: Learning to factorize with transformation for cooperative
  multi-agent reinforcement learning.
\newblock \emph{ICML 2019}, 2019.

\bibitem[Sunehag et~al.(2018)Sunehag, Lever, Gruslys, Czarnecki, Zambaldi,
  Jaderberg, Lanctot, Sonnerat, Leibo, Tuyls, et~al.]{sunehag2018value}
Peter Sunehag, Guy Lever, Audrunas Gruslys, Wojciech~Marian Czarnecki, Vinicius
  Zambaldi, Max Jaderberg, Marc Lanctot, Nicolas Sonnerat, Joel~Z Leibo, Karl
  Tuyls, et~al.
\newblock Value-decomposition networks for cooperative multi-agent learning
  based on team reward.
\newblock In \emph{Proceedings of the 17th International Conference on
  Autonomous Agents and MultiAgent Systems}, pages 2085--2087. International
  Foundation for Autonomous Agents and Multiagent Systems, 2018.

\bibitem[Tan(1993)]{tan1993multi}
Ming Tan.
\newblock Multi-agent reinforcement learning: Independent vs. cooperative
  agents.
\newblock In \emph{Proceedings of the tenth international conference on machine
  learning}, pages 330--337, 1993.

\bibitem[Tang et~al.(2019)Tang, Naphade, Liu, Yang, Birchfield, Wang, Kumar,
  Anastasiu, and Hwang]{tang2019cityflow}
Zheng Tang, Milind Naphade, Ming-Yu Liu, Xiaodong Yang, Stan Birchfield, Shuo
  Wang, Ratnesh Kumar, David Anastasiu, and Jenq-Neng Hwang.
\newblock Cityflow: A city-scale benchmark for multi-target multi-camera
  vehicle tracking and re-identification.
\newblock In \emph{Proceedings of the IEEE Conference on Computer Vision and
  Pattern Recognition}, pages 8797--8806, 2019.

\bibitem[van~den Oord et~al.(2017)van~den Oord, Vinyals, and
  Kavukcuoglu]{van2017neural}
A{\"a}ron van~den Oord, Oriol Vinyals, and Koray Kavukcuoglu.
\newblock Neural discrete representation learning.
\newblock In \emph{NIPS}, 2017.

\bibitem[Vaswani et~al.(2017)Vaswani, Shazeer, Parmar, Uszkoreit, Jones, Gomez,
  Kaiser, and Polosukhin]{vaswani2017attention}
Ashish Vaswani, Noam Shazeer, Niki Parmar, Jakob Uszkoreit, Llion Jones,
  Aidan~N Gomez, {\L}ukasz Kaiser, and Illia Polosukhin.
\newblock Attention is all you need.
\newblock In \emph{Proceedings of the 31st International Conference on Neural
  Information Processing Systems}, pages 6000--6010, 2017.

\bibitem[Wainwright and Jordan(2008)]{wainwright2008graphical}
Martin~J Wainwright and Michael~Irwin Jordan.
\newblock \emph{Graphical models, exponential families, and variational
  inference}.
\newblock Now Publishers Inc, 2008.

\bibitem[Wainwright et~al.(2003)Wainwright, Jaakkola, and
  Willsky]{wainwright2003tree}
Martin~J Wainwright, Tommi~S Jaakkola, and Alan~S Willsky.
\newblock Tree-reweighted belief propagation algorithms and approximate ml
  estimation by pseudo-moment matching.
\newblock In \emph{AISTATS}, 2003.

\bibitem[Wei and Luke(2016)]{wei2016lenient}
Ermo Wei and Sean Luke.
\newblock Lenient learning in independent-learner stochastic cooperative games.
\newblock \emph{The Journal of Machine Learning Research}, 17\penalty0
  (1):\penalty0 2914--2955, 2016.

\bibitem[Wei et~al.(2018)Wei, Wicke, Freelan, and Luke]{wei2018multiagent}
Ermo Wei, Drew Wicke, David Freelan, and Sean Luke.
\newblock Multiagent soft q-learning.
\newblock In \emph{2018 AAAI Spring Symposium Series}, 2018.

\bibitem[Wen et~al.(2018)Wen, Yang, Luo, Wang, and Pan]{wen2018probabilistic}
Ying Wen, Yaodong Yang, Rui Luo, Jun Wang, and Wei Pan.
\newblock Probabilistic recursive reasoning for multi-agent reinforcement
  learning.
\newblock In \emph{International Conference on Learning Representations}, 2018.

\bibitem[Yang et~al.(2018)Yang, Luo, Li, Zhou, Zhang, and Wang]{yang2018mean}
Yaodong Yang, Rui Luo, Minne Li, Ming Zhou, Weinan Zhang, and Jun Wang.
\newblock Mean field multi-agent reinforcement learning.
\newblock \emph{ICML}, 2018.

\bibitem[Yedidia et~al.(2001)Yedidia, Freeman, and Weiss]{yedidia2001bethe}
Jonathan~S Yedidia, William~T Freeman, and Yair Weiss.
\newblock Bethe free energy, kikuchi approximations, and belief propagation
  algorithms.
\newblock \emph{Advances in neural information processing systems}, 13, 2001.

\bibitem[Yuille(2002)]{yuille2002cccp}
Alan~L Yuille.
\newblock Cccp algorithms to minimize the bethe and kikuchi free energies:
  Convergent alternatives to belief propagation.
\newblock \emph{Neural computation}, 14\penalty0 (7):\penalty0 1691--1722,
  2002.

\bibitem[Zhang et~al.(2018)Zhang, Yang, Liu, Zhang, and
  Ba{\c{s}}ar]{zhang2018fully}
Kaiqing Zhang, Zhuoran Yang, Han Liu, Tong Zhang, and Tamer Ba{\c{s}}ar.
\newblock Fully decentralized multi-agent reinforcement learning with networked
  agents.
\newblock \emph{ICML 2018}, 2018.

\bibitem[Zhang et~al.(2020)Zhang, Lin, and Zhang]{zhang2020succinct}
Sai~Qian Zhang, Jieyu Lin, and Qi~Zhang.
\newblock Succinct and robust multi-agent communication with temporal message
  control.
\newblock \emph{Neurips 2020}, 2020.

\bibitem[Zheng et~al.(2018)Zheng, Yang, Cai, Zhou, Zhang, Wang, and
  Yu]{zheng2018magent}
Lianmin Zheng, Jiacheng Yang, Han Cai, Ming Zhou, Weinan Zhang, Jun Wang, and
  Yong Yu.
\newblock Magent: A many-agent reinforcement learning platform for artificial
  collective intelligence.
\newblock In \emph{Thirty-Second AAAI Conference on Artificial Intelligence},
  2018.

\bibitem[Ziebart et~al.(2008)Ziebart, Maas, Bagnell, and
  Dey]{ziebart2008maximum}
Brian~D Ziebart, Andrew Maas, J~Andrew Bagnell, and Anind~K Dey.
\newblock Maximum entropy inverse reinforcement learning.
\newblock 2008.

\end{thebibliography}
\bibliographystyle{plainnat}

\onecolumn
\newpage
\appendix
\title{\bf Appendix: Variational Policy Propagation for Multi-agent Reinforcement Learning  }
\date{}
\author{}
\maketitle

\section{Organization of the Appendix}

In appendix \textbf{\ref{app:more_related_work}}, we discuss more related works and kernel embedding.
In appendix\textbf{ \ref{app:detail_intention_propgation_network}}, we give the parameterization on the variational inference layers. Then we extend the variational policy propagation to other approximations which converges to other solutions of the variational inference. We also provide the equivalent free energy optimization problem for the mean-field approximation.  In Appendix  \textbf{\ref{section:algorithm}}, we provide the pseudo code for the algorithm deferred from the main text. Appendix \textbf{\ref{app:env}} summarizes the configuration of the experiment and MARL environment. Appendix \textbf{\ref{app:setting}} gives more details on baselines and the hyperparameters of neural networks used in our model. Appendix \textbf{\ref{app:ablation}} conducts the ablation study deferred from the main paper. Appendix \textbf{\ref{app:further_exps}} and \textbf{\ref{app:hypers}} give more experimental results and hyperparameters used in the algorithms. At appendix \ref{section:derivation}, we derive the algorithm and prove the proposition \ref{proposition:MRF}.

\section{ Discussions}\label{app:more_related_work}
\subsection{Discuss related work}

The mean-field game (MFG) \citep{huang2006large} generally assumes that agents are identical and interchangeable (it may be extend to several groups of agents.  Agents within a group are identical ). When the number of agents goes to infinity,  MFG can view the state of other agents as a population state distribution. In our work, we do not have such assumptions. The goal of the MFG is to find the Nash equilibrium, while our work aims to the optimal joint policy in the \emph{collaborative} game. At last, MFG assumes that agents are weakly coupled and are mainly affected by the population. We do not have such assumption. In PR2 \citep{wen2018probabilistic}, the joint policy $ \pi(\bm{a}|s)$ is factorized as $\pi(a_i|s)\pi(a_{-i}|s,a_{i})$. The opponent $\pi(a_{-i}|s,a_{i})$ is then approximated  by a neural network $\rho$, which is  optimized by minimizing the KL divergence of probability of trajectory. Once the number of the opponents is large, this opponent model  suffers from the issue of exponentially large action space again. In VPP, the joint policy $\pi$ is factorized as $\pi(\bm{a}|s)\approx \prod q_i(a_i|s)$ (mean-field approximation). Notice there is \textbf{no} exact policy of agent ,i.e.,$\pi_i$. $q_j, j\neq i$ actually approximate the opponent and is optimized by minimizing the KL divergence between $\pi(\bm{a}|s)$ and $\prod q_i(a_i|s)$ using data. Remark that $q_i$ depends on its neighbors (eq3). The advantage of VPP is that we prove the (optimal) policy is MRF. Thus we can leverage the structural information (thus reduce the search space) by the probabilistic inference, while PR2 does not consider such information.  
\subsection{Fixed point with kernel embedding}

We essentially use neural network to approximate the operator $\tilde{\mathcal{T}}$ and kernel embedding $\tilde{\mu}$.  Roughly speaking, once there are enough data, we can believe the learned kernel embedding and learned operator are close enough to the true ones. Therefore the update of  \eqref{equ:output_distribution} and \eqref{equ:message_passing}  would converge to the fixed point of  \eqref{equ:mean_field_solution}. Regarding the generalization error of the deep neural network, unfortunately to the best of our knowledge, it is sill an open problem and therefore we can not give a error bound between learned one and the true one.
For more theoretical guarantee on the kernel embedding, e.g.,  convergence rate on the empirical mean of the kernel embedding please refer to \citep{smola2007hilbert}.

\section{VPP }\label{app:detail_intention_propgation_network}

\subsection{Details on the variational inference layers}

In this section, we give the details on the variational policy propagation network deferred from the main paper. We first illustrate the message passing of the variational inference layers derived in section \ref{section:neural_embedd_vi} and provide the detailed parameterizations.

\begin{figure}[h]
	\centering
	\includegraphics[width=0.8\textwidth]{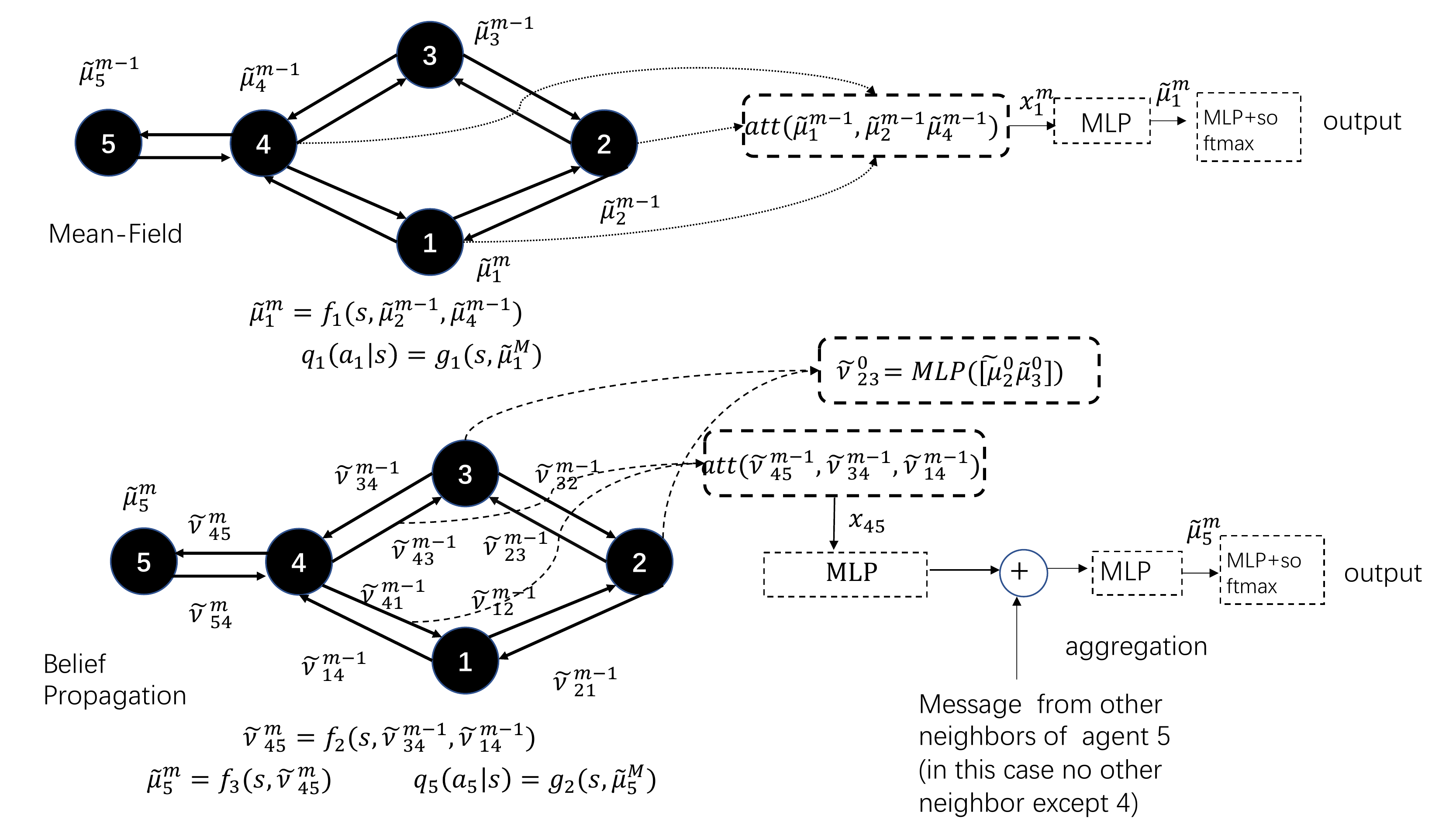}
	\caption{ illustrate the message passing in variational inference layers corresponding to mean-field approximation and loopy belief propagation. The MLP module means the multi-layer perceptron and att module stands for the multi-head attention. }
	\label{fig:intention_propagation}
\end{figure}

The figure \ref{fig:intention_propagation} elaborates the variational inference layers in the Figure \ref{fig:architecture} (b) and (c).  $\Tilde{\mu}_i$ is the  embedding of \emph{policy} of agent $i$. At 0 iteration, every agent makes independent decision $\Tilde{\mu}^0_i$, which is  the embedding of $\psi_i $.  Then agent $i$ sends this embedding to its neighbors. In Figure \ref{fig:intention_propagation}, $\tilde{\mu}_i^{m}$ is the $d$ dimensional embedding of $q_i$ at $m-$th iterations of the rollout in the variational inference layer .  We draw the update rule of $\tilde{\mu}_1^{(m)} $ as example. Agent $1$ receives the embedding  $\tilde{\mu}_2^{m-1},\tilde{\mu}_4^{m-1}$ from its neighbors, and then updates the its own embedding with operator $\tilde{\mathcal{T}}$ which is approximated by a neural network $f_1$. After $M$ iterations, we obtain $\tilde{\mu}_1^{M}$ and output the policy distribution $q_1$ using \eqref{equ:output_distribution}. Similar procedure holds for other agents. At each RL step $t$, we do this procedure (with M iterations) once, which  basically unrolls the fixed point update rule,  to generate joint policy. In practice $M$  is small, e.g., $M=2$ or $3$.

Recall that  in the mean-field approximation we just use the embedding of $\psi_i$, while In the loopy belief propagation, we also need the embedding of $\psi_{ij}$ which is $\tilde{v}_{ij}^{0}$ at the initial step. One simple way to construct $\tilde{v}_{ij}^{0}$ is to  concatenate  the information of the node $i$ and $j$ and map them to $\tilde{v}_{ij}^{0}$ with a neural network. Then we follow the fixed point update rule in the loopy belief propagation to generate a message passing rule. E.g., the message $\tilde{\nu}_{45}$ from agent 4 to agent 5 is a function of message $\tilde{\nu}_{34}$ and $\tilde{\nu}_{14}$. At the final round $M$, agent 5 aggregates all messages from its neighbors (in this figure, it is just agent 4) and then transforms it into its own (policy) embedding $\tilde{\mu}_5^M$ on the node. Then he makes the decision $a_5$ sampled from $q_5$.

\begin{figure}
	\centering
	\includegraphics[width=0.6\textwidth]{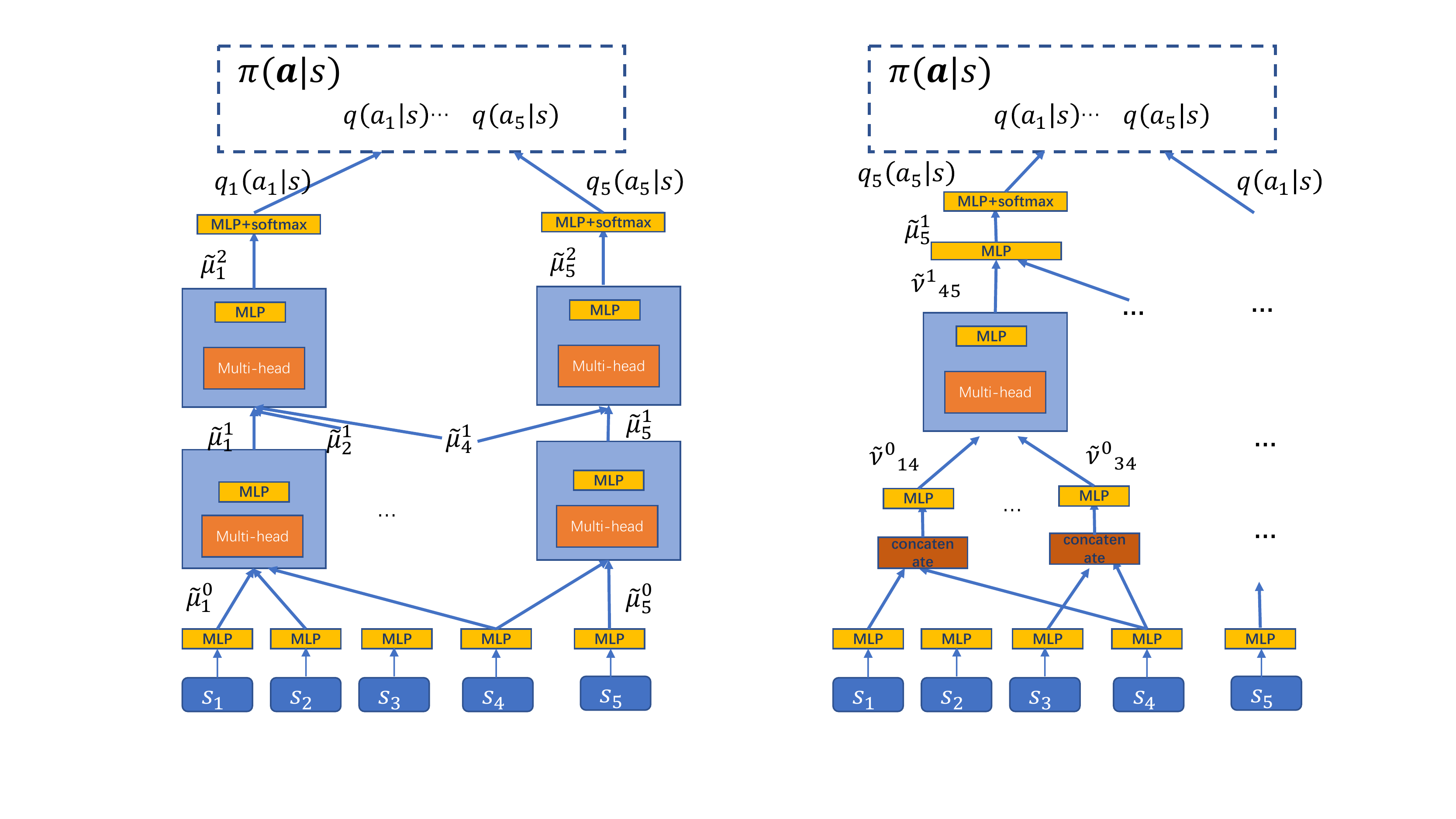}
	\caption{ Details of the neural network corresponding to mean-field approximation is presented in the left panel. The right one corresponds to loopy belief propagation.  }
	\label{fig:network_parameterization}
\end{figure}

We illustrate the parameterization of graph neural network in Figure \ref{fig:network_parameterization}. If the action space is discrete, the output $q_i(a_i|s)$ is a softmax function. When it is continuous, we can output a Gaussian distribution (mean and variance) with the reparametrization trick \citep{kingma2019introduction}. In the left panel of Figure \ref{fig:network_parameterization}, we draw  2 round message passing of the mean-field approximation.  Each agent observe its own state $s_i$. After a MLP layer, we get a embedding $\tilde{\mu}^0_i$, which is the initial embedding of the policy. Here the weight can be shared across agent or not. It depends on whether agents are identical or not. In the following,  we use agent 1 as an example.  Agent 1 receives the embedding $\tilde{\mu}^0_2$ and $\tilde{\mu}^0_4$ from its neighbor. After a multi-head attention (over the neighbors) and MLP module, we obtain  new embedding $\tilde{\mu}^1_1$ of agent 1.  In particular, for  agent $i$  and head $l$, we have 
$a_{i,j} = \frac{\exp\big( W^l_Q\tilde{\mu}_i^{m-1} (W^l_K\tilde{\mu}_j^{m-1})^T\big)}{\sum_{k\in \mathcal{N}_i} \exp\big( W^l_Q\tilde{\mu}_i^{m-1} (W^l_K\tilde{\mu}_k^{m-1})^T\big)}.$  Then the embedding of agent $i$ in the next round is $ \tilde{\mu}_i^{m}=\sigma(W_s s_i^{m-1}+\sum_{j\in\mathcal{N}_i}a_{i,j} W^l_V\tilde{\mu}_j^{m-1}  ).$ Notice that we also do message passing for state which is $s_i^{m}= \sigma(W_1 s_i^{m-1} +W_2\sum_{j\in \mathcal{N}_i}s_j^{m-1})$.
We do this message-passing procedure with two rounds for agent 1 and obtain the final result $\tilde{\mu}_1^2$.   Then the embedding $\tilde{\mu}_1^2$ passes a MLP+softmax layer and outputs probability of action, i.e. $q_1(a_1|s)$.  

In the loopy belief propagation, the parameterization has a similar spirit but now we need to generate the embedding  of the message on the edge $\tilde{\nu}_{ij}$ besides the  embedding $\tilde{\mu}_i$ on the node. To generate the initial embedding $\tilde{\nu}_{ij}^0$, we first generate the embedding of the node $i$ and $j$, i.e., $ \tilde{\mu}_i^0$ and $\tilde{\mu}_j^0$ . Then we concatenate  the information of them in the following way $\tilde{\nu}_{ij}^0 = MLP([\tilde{\mu}_i^0,\tilde{\mu}_j^0])$. We apply the multi-head attention +MLP module on the $\tilde{\nu}_{ij}$ again. But notice that the attention $a_{ij}$ now is over the corresponding edge rather than the nodes.  Then we have $\tilde{\nu}_{ij}^m = \sigma(W_s [s_i^{m-1},s_j^{m-1}] + \sum_{k\in \mathcal{N}_i \backslash j} a_{ik}\tilde{\nu}_{ki}^{m-1})$. At the final the stage, we aggregate the message  $\tilde{\nu}_{ki}^m$ on the edge to generate the embedding on the node $\tilde{\mu}_i^m$, which is $\tilde{\mu}_i^m = \sigma(W_3s^{m}_i +W_4 \sum_{k\in \mathcal{N}_i}\tilde{\nu}_{ki}^m)$.

\subsection{Extension to other variational inference methods and Neural networks}\label{section:other_variational_inference}

In this section, we show how to approximate the joint policy with the Loopy Belief Propagation in the variational inference \citep{yedidia2001bethe}. This will  lead to a new form of neural networks beyond vanilla mean-field approximation.   

The objective function in Loop Belief Propagation is the Beth Free energy \citep{yedidia2001bethe}. Different from the mean-field approximation, it introduces another variational variable $q_{ij}$, which brings more flexibility on the approximation. The following is the corresponding optimization problem.

\begin{flalign}
\begin{aligned}
\min_{ q_i,q_{ij}, \{i,j\}\in \mathcal{E}}-\sum_i(|\mathcal{N}_i|-1) &\int q_i(a_i|s)\log\frac{q_i(a_i|s)}{\psi_i(s,a_i)} da_i\\
&+\sum_{ij}\int q_{ij}(a_i,a_j|s)\log \frac{q_{ij}(a_i,a_j|s)}{\psi_{ij}(s,a_i,a_j)\psi_i(s,a_i)\psi_j(s,a_j)}da_ida_j.\\
&s.t. \int q_{ij}(a_i,a_j|s)da_j=q_i(a_j|s), \int q_{ij}(a_i,a_j|s)da_i=q_j(a_j|s)
\end{aligned}
\end{flalign}

Solve above problem, we have the fixed point algorithm
$$ m_{ij}(a_j|s)\leftarrow \int \prod_{k\in \mathcal{N}_i\setminus j } m_{ki}(a_i|s)\psi_i(s,a_i)\psi_{ij}(s,a_i,a_j)da_i, $$
$$q_i(a_i|s)\leftarrow \psi_i(s,a_i)\prod_{j\in \mathcal{N}_i} m_{ji}(a_i|s).$$

Similar to the mean-field approximation case, we have
$$m_{ij}(a_j|s)= f(a_j,s, \{m_{ki}\}_{k\in \mathcal{N}_i \setminus j} ),$$
$$q_i(a_i|s) = g(a_i,s,\{m_{ki}\}_{k\in \mathcal{N}_i}), $$

It says the message $m_{ij}$ and marginals $q_i$ are functionals of messages from neighbors.

Denote the embedding $\tilde{\nu}_{ij}=\int \psi_j(s,a_j)m_{ij}(a_j|s) da_j$ and $\tilde{\mu}_i=\int \psi_i(s,a_i)q_i(a_i|s)da_i,$ we have 

$\tilde{\nu}_{ij} =\tilde{\mathcal{T}}_1\circ \big( s, \{ \tilde{\nu}_{ki}\}_{k\in \mathcal{N}_i \setminus j}  \big)$,
$ \tilde{\mu}_i = \tilde{\mathcal{T}}_2\circ \big( s, \{\tilde{\nu}_{ki}\}_{k\in \mathcal{N}_i} \big).$

Again, we can parameterize above equation by  neural networks, e.g., a simple way is  $\tilde{\nu}_{ij} =\sigma\big(  W_1s+ W_2 \sum_{k\in \mathcal{N}_i \setminus j} \tilde{\nu}_{ki} \big),$ $ \tilde{\mu}_i= \sigma \big( W_3s+W_4 \sum_{k\in \mathcal{N}_i} \tilde{\nu}_{ki} \big).$

Following similar way, we can derive different variational policy propagation algorithms by changing different objective function which corresponds to e.g., double-loop belief propagation\citep{yuille2002cccp}, tree-reweighted belief propagation \citep{wainwright2003tree} and many others.

\subsection{Free energy form of mean-field approximation}\label{app:free_energy_mean}

According to \cite{wainwright2008graphical}, the mean-field approximation actually minimize following free energy.

\begin{equation}
\min_{q_i}\int \prod_i q_i(a_i|s) \big( \sum_i \log q_i(a_i|s) - \log \pi(\mathbf{a}|s) \big) d\mathbf{a}.\quad s.t. \int q(a_i|s) da_i = 1
\end{equation}

\section{Algorithm}\label{section:algorithm}

\begin{algorithm}[h]
	\caption{variational policy propagation}\label{alg:IP}
	\begin{algorithmic}
		\STATE{\textbf{Inputs}: Replay buffer $D$. $V_i$, $Q_i$ for each agent $i$. The policy $\pi_\theta$ with outputs $\{q_{i,\theta}\}_{i=1}^{N}$. Learning rate $l_\eta,l_{\kappa,} l_\theta$. Moving average parameter $\tau$ for the target network} \\
		\FOR{each iteration}
		\FOR{each environment step}
		\STATE sample $\mathbf{a_t}\sim \prod q_{i,\theta}(a_i^t|s^t)$ from the  variational inference  layer.
		sample the next state from the environment $s^{t+1}\sim p(s^{t+1}|s^t,\mathbf{a^t})$. Aggregate the data into replay buffer $D\leftarrow D \bigcup \big( s_i^t, a_i^t, r_i^t, s_i^{t+1} \big)_{i=1}^N$ 
		\ENDFOR
		\FOR{each gradient step}
		\STATE{}
		update the parameter of value function, state-action value function, policy and target network, i.e.,  $\eta_i$, $\kappa_i$, $\theta$, $\Bar{\eta}_i$ using following way
		$$\eta_i\leftarrow\eta_i-l_\eta\nabla J_V(\eta_i),\kappa_i\leftarrow\kappa_i-l_\kappa\nabla J_Q(\kappa_i) $$
		$$\theta\leftarrow\theta-l_\theta\nabla J(\theta),\Bar{\eta}_i\leftarrow\tau \eta_i+(1-\tau)\Bar{\eta}_i$$
		\ENDFOR
		\ENDFOR
	\end{algorithmic}
\end{algorithm}

Remark: To calculate the loss function $J(\eta_i)$, each agent needs to sample the global state and $(a_i,a_{\mathcal{N}_i})$. Thus we first sample a global state from the replay buffer and then sample all action $\mathbf{a}$ once using the variational inference layer once.

\section{Further details about environments and experimental setting}
\label{app:env}




Table \ref{tb:setting} summarizes the setting of the tasks in our experiment.
\begin{table}[ht!]
	
	\begin{center}
		\caption{\small Tasks. We evaluate MARL algorithms on more than 10 different tasks from three different environments.}
		\label{tb:setting}
		
		\small
		\setlength{\tabcolsep}{2pt}
		\begin{tabular}{c | c| c  }
			\hline
			Env & Scenarios & \text{\#}agents (N)\\
			\hline
			\multirow{2}{*}{CityFlow} 
			& Realworld:Hang Zhou & N=16 \\
			& Realworld:Manhattan & N=96 \\
			& Synthetic Map & N=49, 100, 225, 1225 \\
			\hline   
			\multirow{4}{*}{MPE} 
			& Cooperative Nav. & N=15, 30, 200 \\
			& Heterogeneous Nav. & N=100 \\
			& Cooperative Push  & N=100   \\
			& Prey and Predator  & N=100 \\
			\hline   
			MAgent & Jungle & N=20, F=12  \\
			\hline
			
		\end{tabular}
	\end{center}
	\vspace{-3mm}
\end{table}

\subsection{CityFlow}


CityFlow~\citep{tang2019cityflow} is an open-source MARL environment for large-scale city traffic signal control~\footnote{\url{https://github.com/cityflow-project/CityFlow}}. 
After the traffic road map and flow data being fed into the simulators, each vehicle moves from its origin location to the destination. The traffic data contains bidirectional and dynamic flows with turning traffic.
We evaluate different methods on both real-world and synthetic traffic data.
For real-world data, we select traffic flow data from Gudang sub-district, Hangzhou, China  and Manhattan, USA ~\footnote{We download the maps from \url{https://github.com/traffic-signal-control/sample-code}.}.
For synthetic data, we simulate several different road networks: $7\times7$ grid network ($N=49$) and  large-scale grid networks with  $N=10\times10=100$ , $15\times 15=225$, $35\times35=1225$. Each traffic light at the intersection is the agent. In the real-world setting (Hang Zhou, Manhattan), the graph is a 2-d grid induced by the roadmap. Particularly,  the roads are edges which connect the node (agent) of the graph. For the synthetic data, the map is a $n*n$ 2-d grid (Something like	Figure \ref{fig:toy_example}), where edges represents road, node is the traffic light. We present the more experimental results deferred from the main paper in Figure \ref{fig:exp_futher_cityflow}.

\subsection{MPE}

In MPE~\citep{mordatch2017emergence}~\footnote{To make the environment more computation-efficient, \citet{liu2019pic} provided an improved version of MPE. The code are released in~\url{https://github.com/IouJenLiu/PIC}.}, the observation of each agent contains relative location and velocity of neighboring agents and landmarks.
The number of visible neighbors in an agent’s observation is equal to or less than 10. 
In some scenarios, the observation may contain relative location and velocity of neighboring agents and landmarks.

We consider four scenarios in MPE. 
(1) \textit{cooperative navigation}: $N$ agents work together and move to cover $L$ landmarks. If these agents get closer to landmarks, they will obtain a larger reward. 
In this scenario, the agent observes its location and velocity, and the relative location of the nearest $5$ landmarks and $N$ agents. The observation dimension is 26.
(2) \textit{prey and predator}: $N$ slower cooperating agents must chase the faster adversaries around a randomly generated environment with $L$ large landmarks. Note that, the landmarks impede the way of all agents and adversaries. This property makes the scenario much more challenging. 
In this scenario, the agent observes its location and velocity, and the relative location of the nearest $5$ landmarks and $5$ preys. The observation dimension is 34.
(3) \textit{cooperative push}: $N$ cooperating agents are rewarded to push a large ball to a landmark. 
In this scenario, each agent can observe $10$ nearest agents and $5$ nearest landmarks.
The observation dimension is 28.
(4) \textit{heterogeneous navigation}: this scenario is similar with cooperative navigation except dividing $N$ agents into $\frac{N}{2}$ big and slow agents and $\frac{N}{2}$ small and fast agents. If small agents collide with big agents, they will obtain a large negative reward.
In this scenario, each agent can observe $10$ nearest agents and $5$ nearest landmarks.
The observation dimension is 26.

Further details about this environment can be found at~\url{https://github.com/IouJenLiu/PIC}.

\subsection{MAgent}

MAgent~\citep{zheng2018magent} is a grid-world platform and serves another popular environment platform for evaluating MARL algorithms.
\citet{jiang2020graph} tested their method on two scenarios: \textit{jungle} and \textit{battle}.
In \textit{jungle}, there are $N$ agents and $F$ foods. The agents are rewarded by positive reward if they eat food, but gets higher
reward if they attack other agents. 
This is an interesting scenario, which is called by \textit{moral dilemma}.
In \textit{battle}, $N$ agents learn to fight against several enemies, which is very similar with the \textit{prey and predator} scenario in MPE.
In our experiment, we evaluate our methods on \textit{jungle}.

In our experiment, the size for the grid-world environment is $30\times30$.
Each agent refers to one grid and can observe $11\times11$ grids centered at the agent and its own coordinates.
The actions includes moving and attacking along the coordinates. Further details about this environment can be found at~\url{https://github.com/geek-ai/MAgent} and~\url{https://github.com/PKU-AI-Edge/DGN}.

\subsection{Computation Resource }

We run all experiment with two tesla v100 GPUs.

\section{Further Details on Settings} \label{app:setting}

\subsection{Further description of our baselines}
\label{app:baseline}


In homogeneous settings, the input to the centralized critic in MADDPG is the concatenation of all agent's observations and actions along the specified agent order, which doesn't hold the property of \textit{permutation invariance}. We follow the similar setting in~\citep{liu2019pic} and shuffle the agents' observations and actions in training batch~\footnote{This operation doesn't change the state of the actions.}. In COMA \citep{foerster2018counterfactual}, it directly assume the poilcy is factorized. It calculates the counterfactual baseline  to address the credit assignment problem in MARL. In our experiment, since we can observe each reward function, each agent can directly approximate the Q function without counterfactual baseline. MFQ  derives the algorithm  with the mean action of the neighbor \citep{yang2018mean}.  It needs the assumption that agents are identical. \citet{chu2020multi} provide the algorithm Neurcomm, which assumes the spatial markov property of the problem.

\section{Ablation Studies}
\label{app:ablation}

\subsection{Independent policy vs VPP }

\begin{figure*}
	\centering
	\includegraphics[width=.4\textwidth]{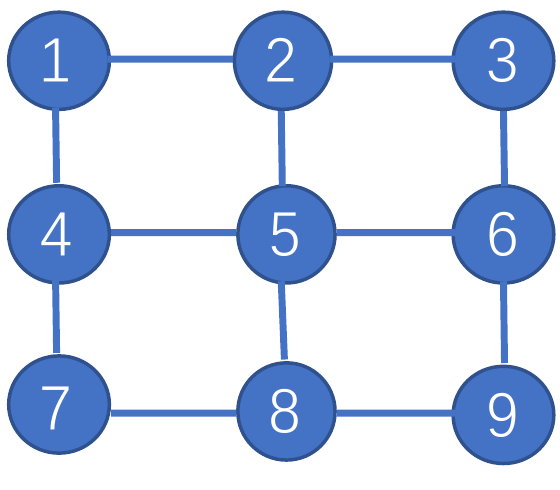}\quad
	\includegraphics[width=.45\textwidth]{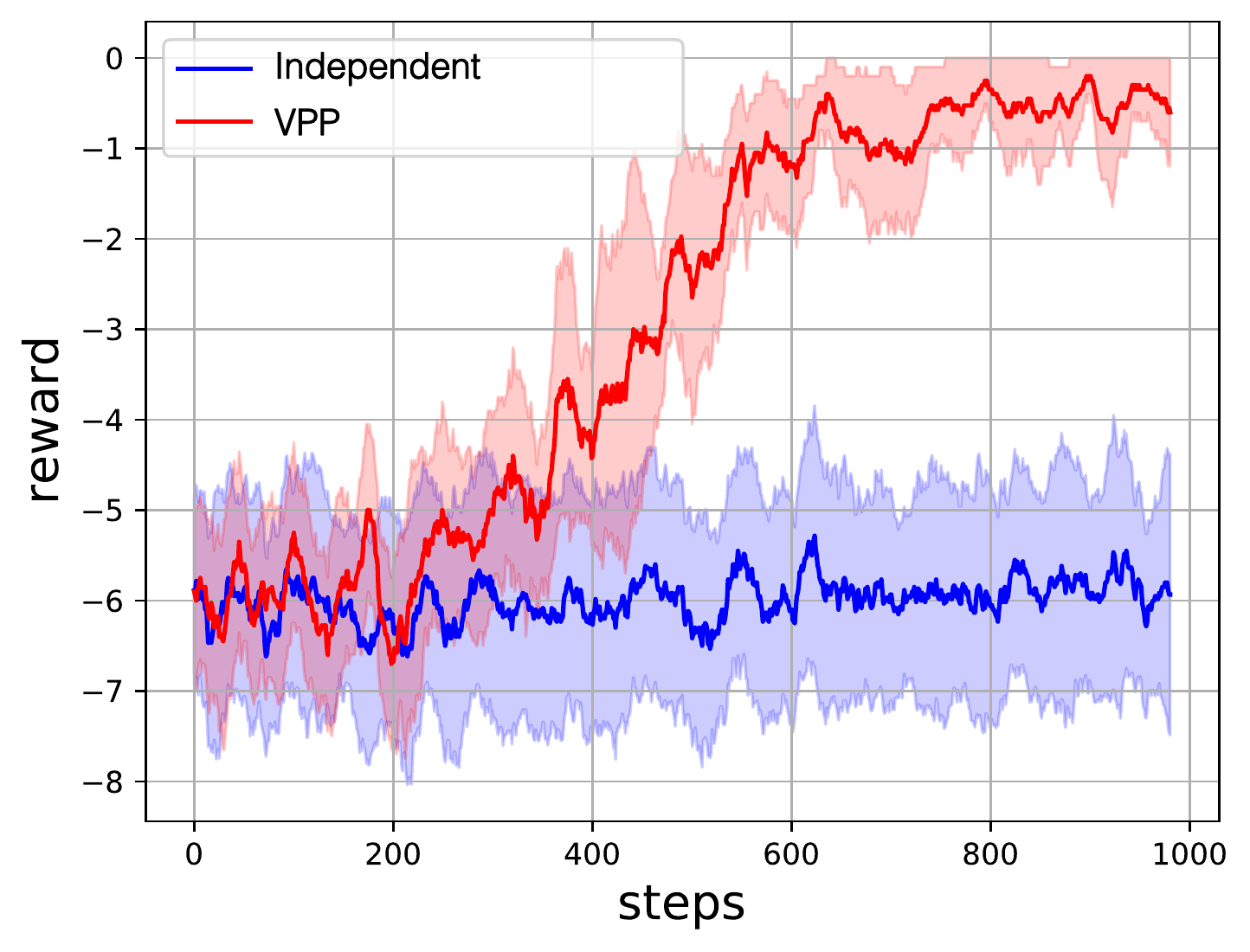}\quad
	\caption{(a) a toy task on 2d-grid. (b) The performance of independent policy and variational policy propagation.
	}
	\label{fig:toy_example}
\end{figure*}

We first give a toy example where the independent policy (without communication) fails. To implement such algorithm, we just replace the variational policy propagation network by a independent policy network and remain other parts the same.   Think about a $3\times3$ 2d-grid in Figure \ref{fig:toy_example}  where the  global state (can be observed by all agents) is a constant scalar (thus no information). Each agent chooses an action $a_i$ = 0 or 1. The aim is to maximize a reward $-(a_1-a_2 )^2- (a_1-a_4)^2-(a_2-a_3)^2-...-(a_8-a_9)^2$, (i.e., summation of the reward function on edges). Obviously the optimal value is 0. The optimal policy for agents is $a_1=a_2=,...,a_9=0$ or $a_1=a_2=,...,a_9=1$. However independent policy fails, since each agent does not know how its allies pick the action. Thus the learned policy is random. We show the result of this toy example in Figure \ref{fig:toy_example}, where variational policy propagation learns optimal policy.

\subsection{Graph types, number of neighbors,  hop size (rounds of message passing), k-nearest neighbor vs. random graph, time interval to update Adjacency Matrix}

\begin{figure*}
	\centering
	\subfloat[CityFlow:7*7]{\includegraphics[width=.4\textwidth]{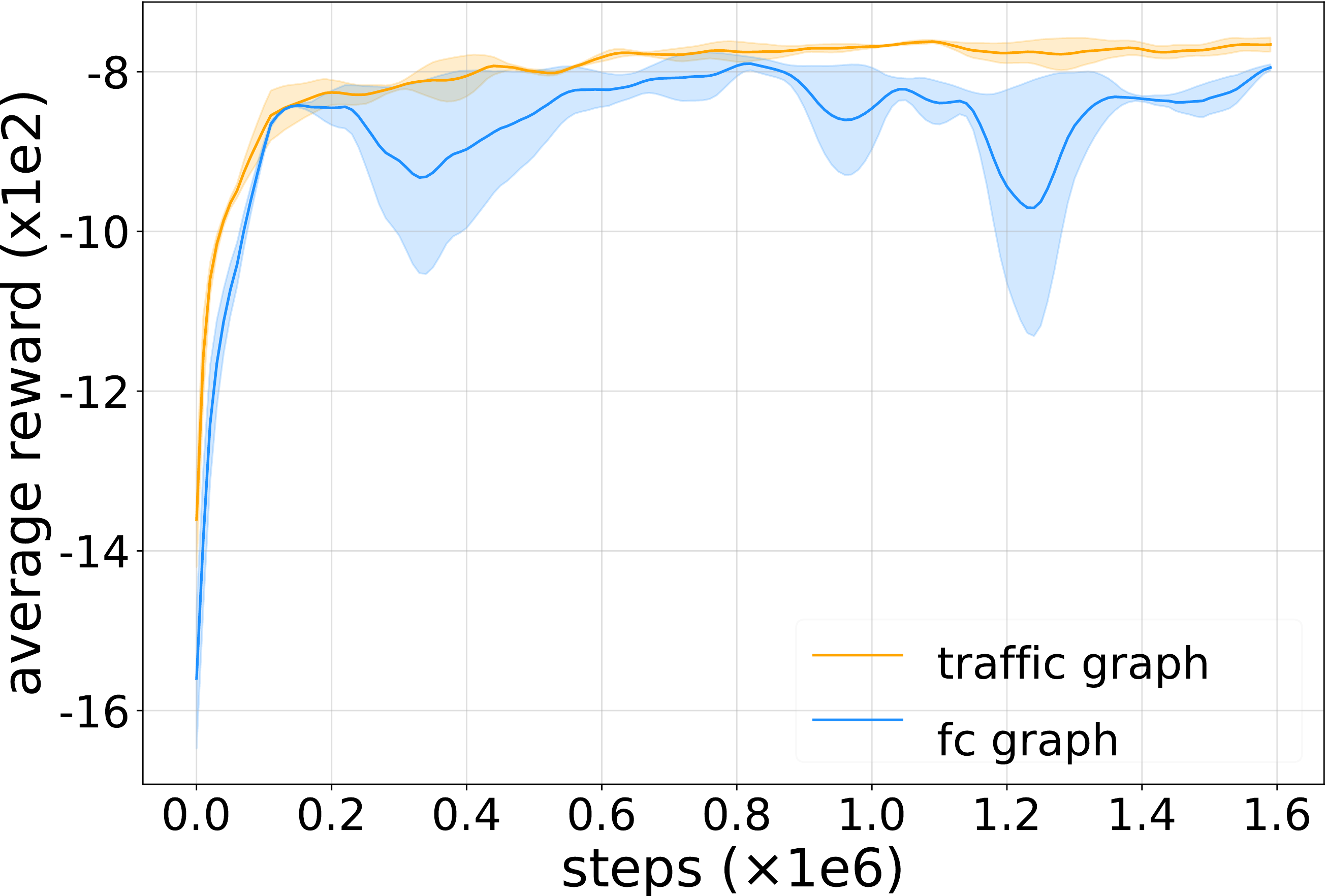}}\quad
	\subfloat[CityFlow:10*10]{\includegraphics[width=.4\textwidth]{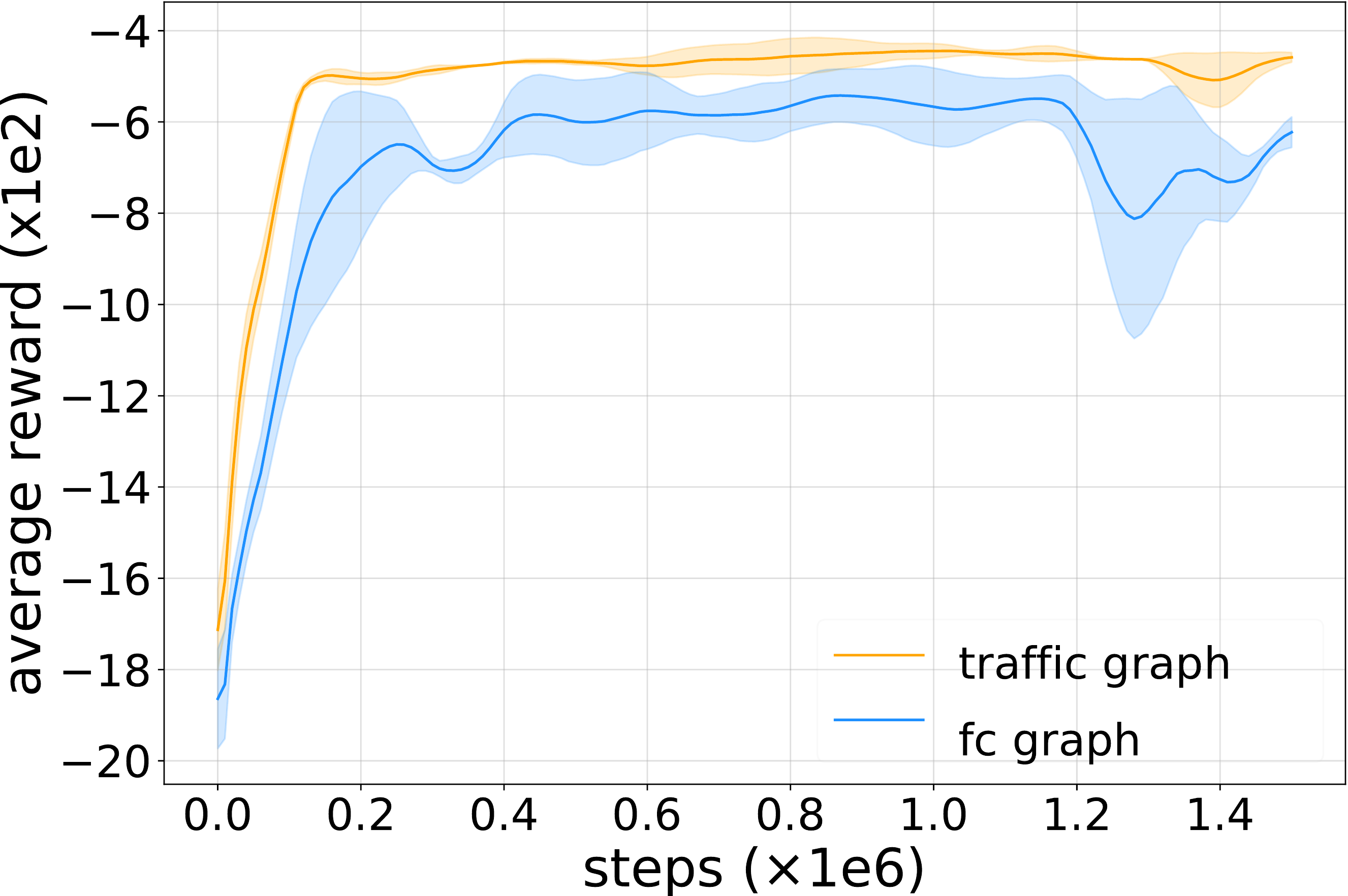}}\quad
	\subfloat[neighbors on Cooperative Nav. (N=30)]{\includegraphics[width=.4\textwidth]{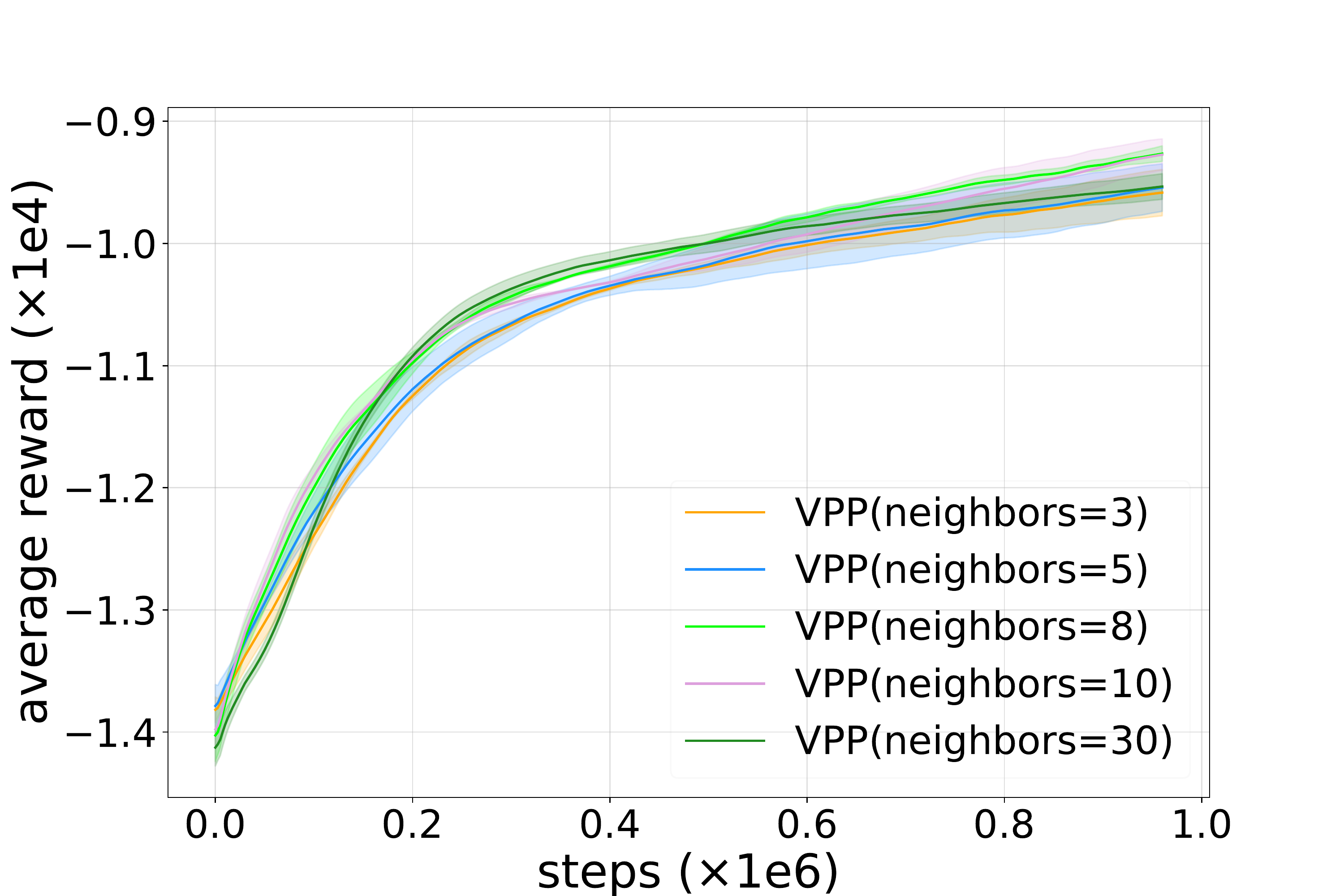}}\quad
	\subfloat[hop on Cooperative Nav. (N=30)]{\includegraphics[width=.4\textwidth]{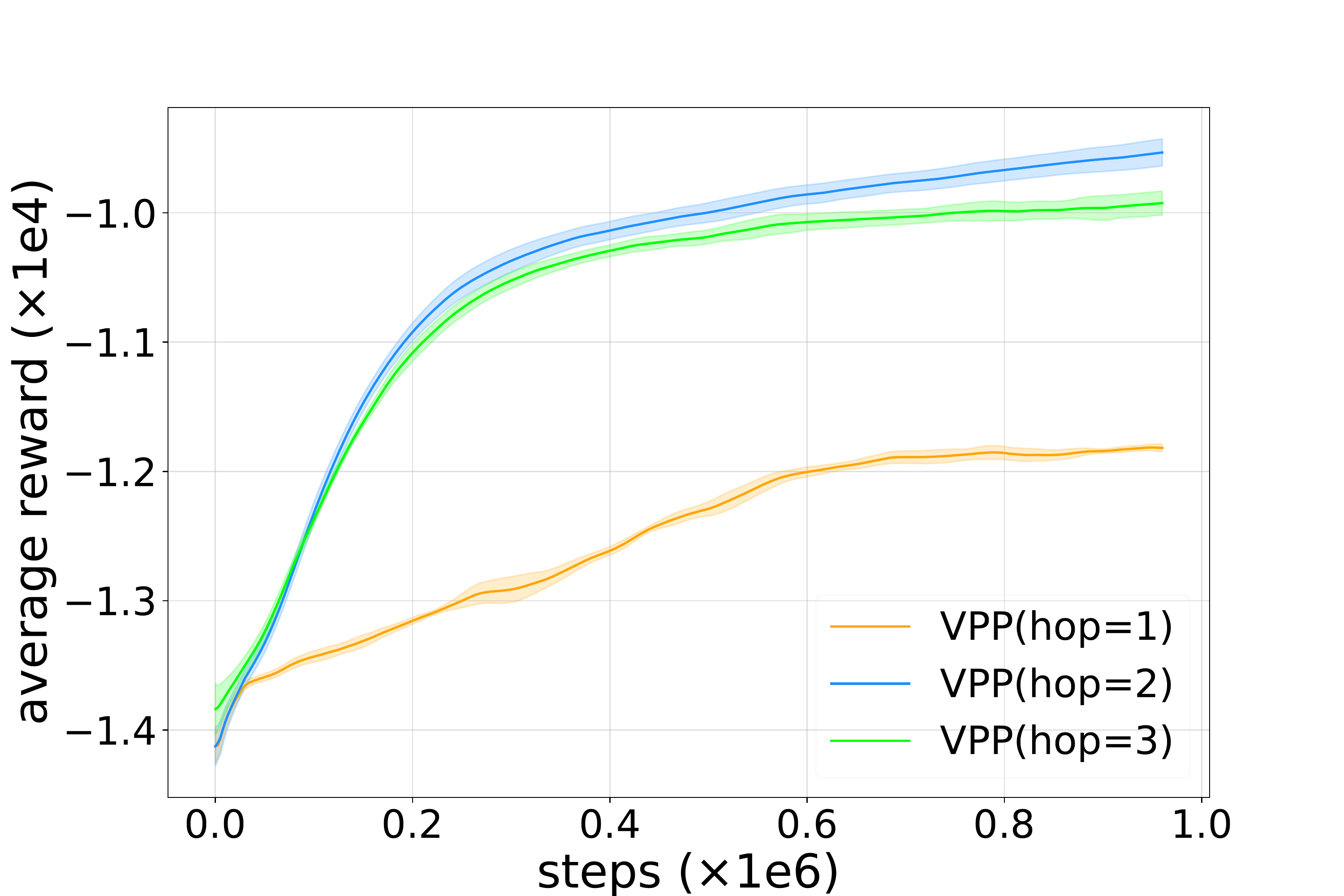}}\quad
	\subfloat[Random graph on Cooperative Nav. (N=30)]{\includegraphics[width=.4\textwidth]{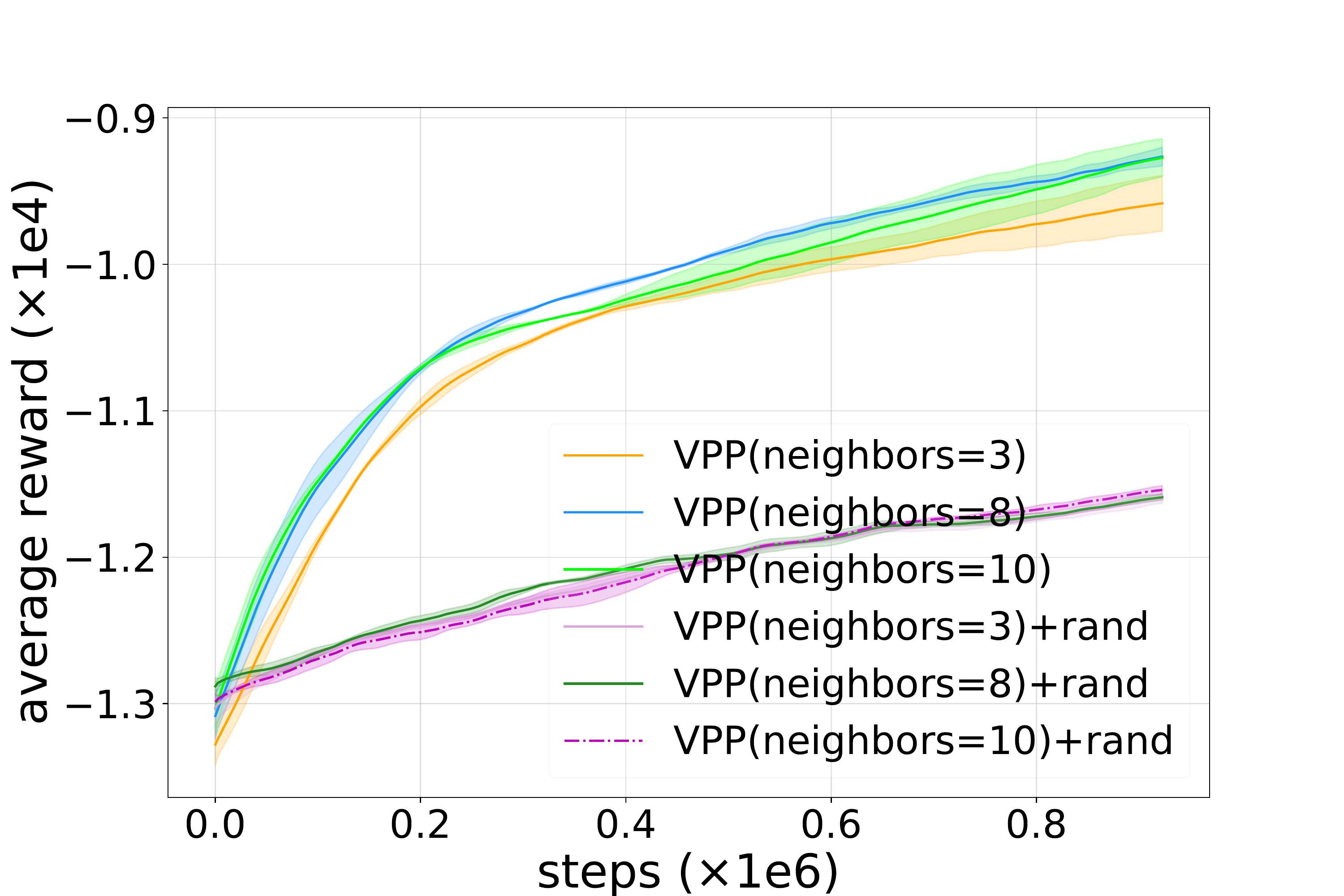}}\quad
	\subfloat[Update adjacency matrix every n=1,5,10 steps.  Cooperative Nav. (N=30)]{\includegraphics[width=.4\textwidth]{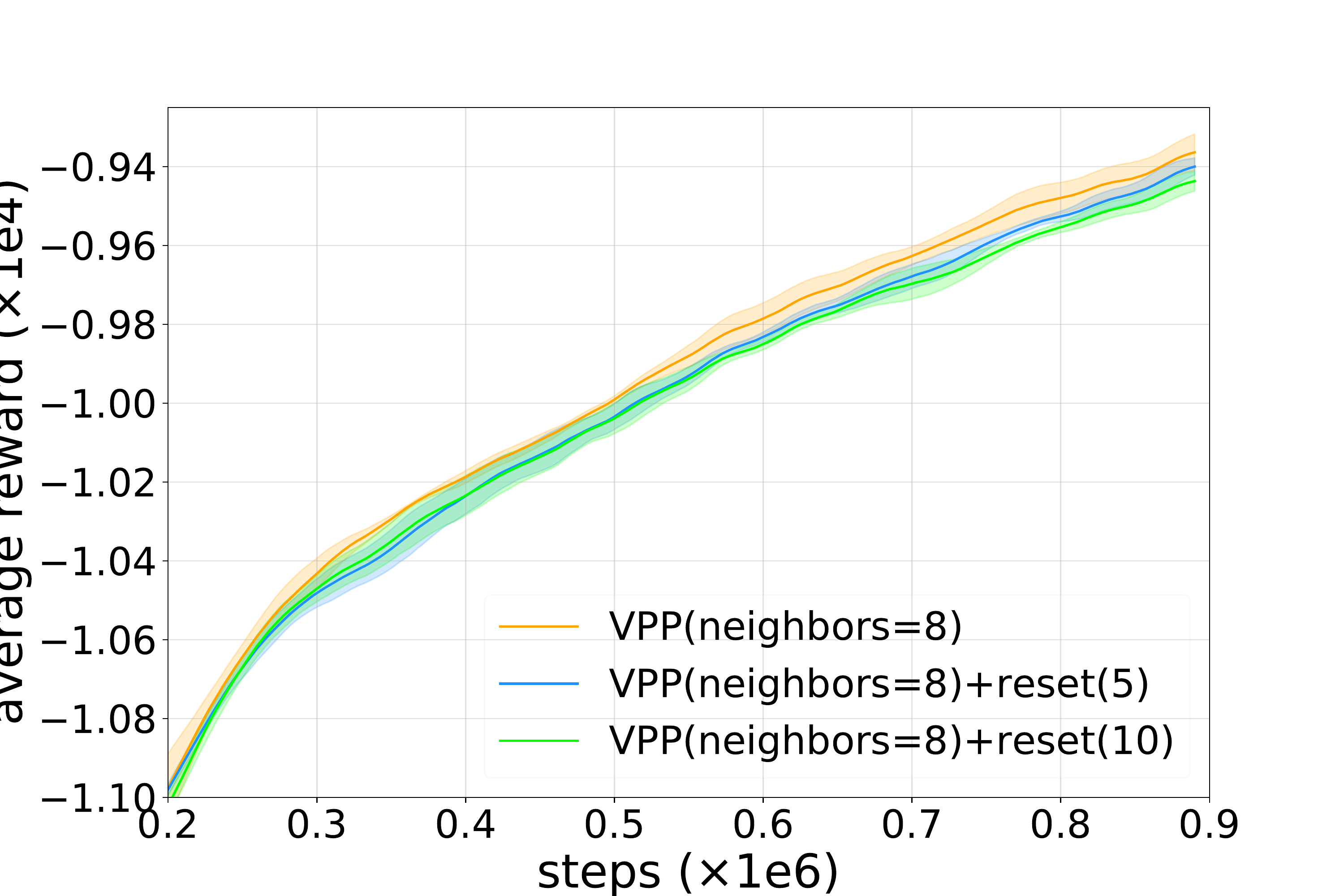}}\quad
	\caption{Performance of the proposed method based on different ablation settings. 
		(a) Traffic graph vs fully connected~(fc) graph on CityFlow (N=49).
		(b) Traffic graph vs fully connected~(fc) graph on CityFlow (N=100).
		(c) Cooperative Nav. (N=30): Different number of neighbors.
		(d) Cooperative Nav. (N=30): Different hop size of message passing.
		(e) Cooperative Nav. (N=30): Construct random graph vs $k$-nearest-neighbor graph ($k=3, 8, 10$).
		(f) Cooperative Nav. (N=30): Update $8$-nearest-neighbor graph every n environment steps (1, 5 and 10 respectively).
	}
	\label{fig:exp_ablation}
\end{figure*}

We conduct a set of ablation studies related to graph types, the number of neighbors,  hop size, k-nearest neighbor vs. random graph, time interval to update adjacency matrix.
Figure~\ref{fig:exp_ablation}(a) and Figure~\ref{fig:exp_ablation}(b) demonstrate the performance of our method on traffic graph and fully-connected graph on the scenarios (N=49 and N=100) of CityFlow. In the experiment, each agent can only get the information from its neighbors through message passing (state embedding and the policy embedding). The result makes sense, since the traffic graph represents the structure of the map. Although the agent in the  fully connected graph would obtain global information, it may introduce irrelevant information from agents far away.

Figure~\ref{fig:exp_ablation}(c) and Figure~\ref{fig:exp_ablation}(d) demonstrate the  performance under different number of neighbors and hop size (round of message passing)  on \textit{cooperative navigation} (N=30) respectively. The algorithm with neighbors=8 has the best performance. Again the fully connected graph (neighbors=30) may introduce the irrelevant information of the agents in distance. Thus its performance is worse than the algorithm with graph constructed by the K-nearest neighbor. In addition, the fully connected graph  introduces more computations in the training. In Figure~\ref{fig:exp_ablation}(d), we increase the hop-size from 1 to 3. The performance of VPP with hop=2 is much better than that with hop=1. While VPP with hop=3 is just slightly better than that with hop=2. It means graph neural network with hop size =2 has aggregated enough information.

In Figure~\ref{fig:exp_ablation}(e), we test the importance of the k-nearest neighbor structure. VPP(neighbors=3)  + random means that we  pick $3$ agents uniformly at random as the neighbors. Obviously, VPP with K-nearest neighbors outperforms the  VPP with random graph a lot.  In Figure~\ref{fig:exp_ablation}(f), we update adjacency matrix every 1, 5, 10 steps. VPP(neighbors=8) denotes that we update the adjacency matrix every step, while VPP(neighbors=8)+reset(5) and VPP(neighbors=8)+reset(10) means that we update adjacency matrix every 5 and 10 steps respectively. Obviously, VPP(neighbors=8) has the best result. VPP(neighbors=8)  +reset(5) is better than VPP(neighbors=8)   +reset(10). The result makes sense, since the adjacency matrix is more accurate if the update interval is smaller.

\subsection{Assumption Violation}
\begin{figure}
	\centering
	\subfloat[Coop. Nav. Violation (N=30)]{\includegraphics[width=.5\textwidth]{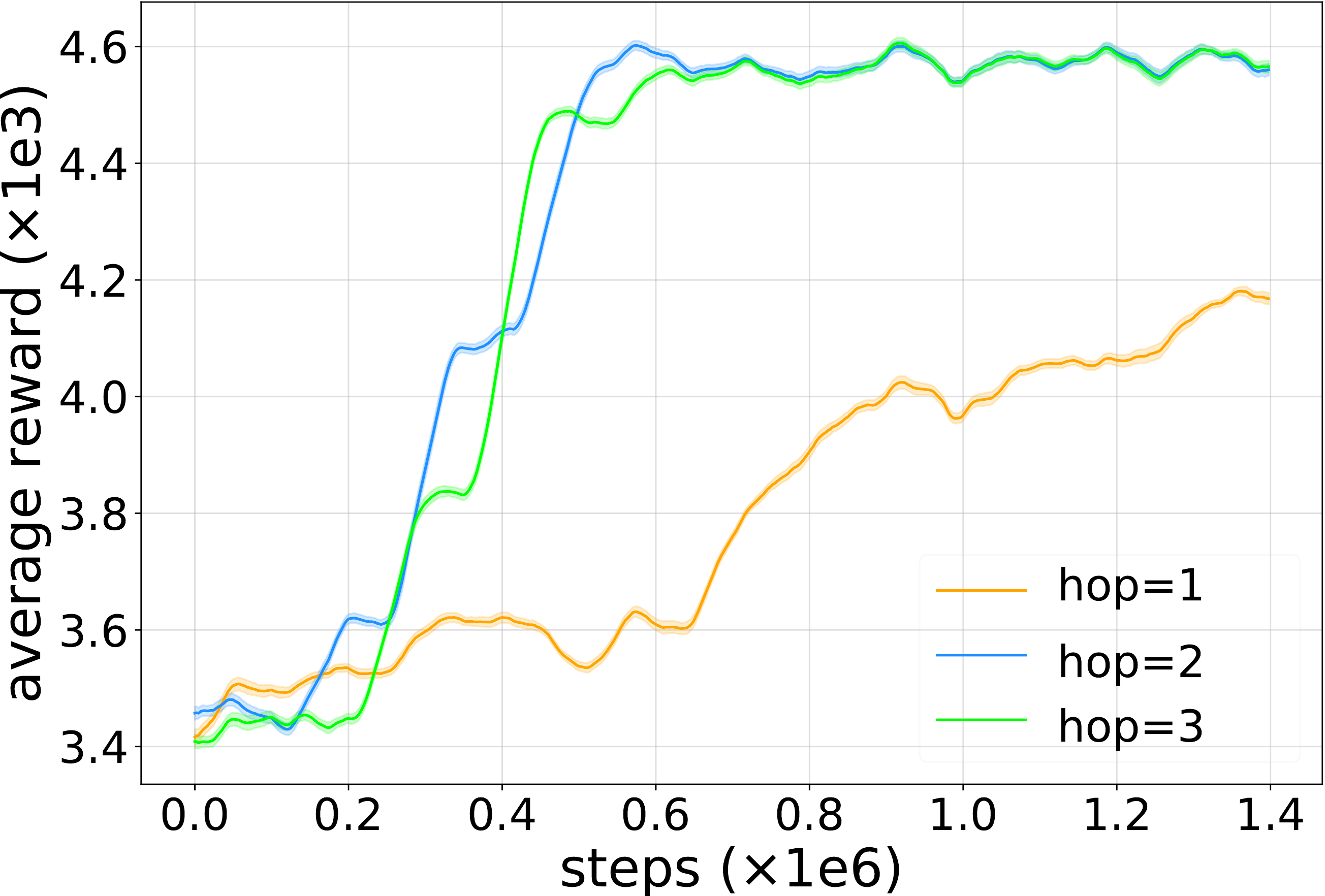}}\quad
	\caption{Further experimental results. 
		Cooperative navigation (N=30) with assumption violation.
	}
	\label{fig:exp_addition}
\end{figure}

The aforementioned experimental evaluations are based on the mild assumption: the actions of other agents in distance   would not affect the learner because of their physical distance (the distance can be defined in many ways other than Euclidean distance). It would be interesting to see the performance where the assumption is violated. To this end, we modify the reward in the experiment of cooperative navigation. In particular, the reward is defined by $r=r_1+r_2$, where $r_1$ encourages the agents to cover (get close to) landmarks and $r_2$ is the log function of the distances between agents (farther agents have larger impact). To make a violation, we let $r_2$ dominate the reward. We conduct the experiments with $hop=1, 2, 3$. Figure~\ref{fig:exp_addition} shows that the rewards obtained by our methods are $4115\pm21$, $4564\pm22$, and $4586\pm25$ respectively. It’s expected in this scenario, since we should use large hop to collect information from the far-away agents.

\section{Further experimental results}
\label{app:further_exps}

\begin{figure*}
	\centering
	\subfloat[CityFlow:Hang Zhou 4*4]{\includegraphics[width=.31\textwidth]{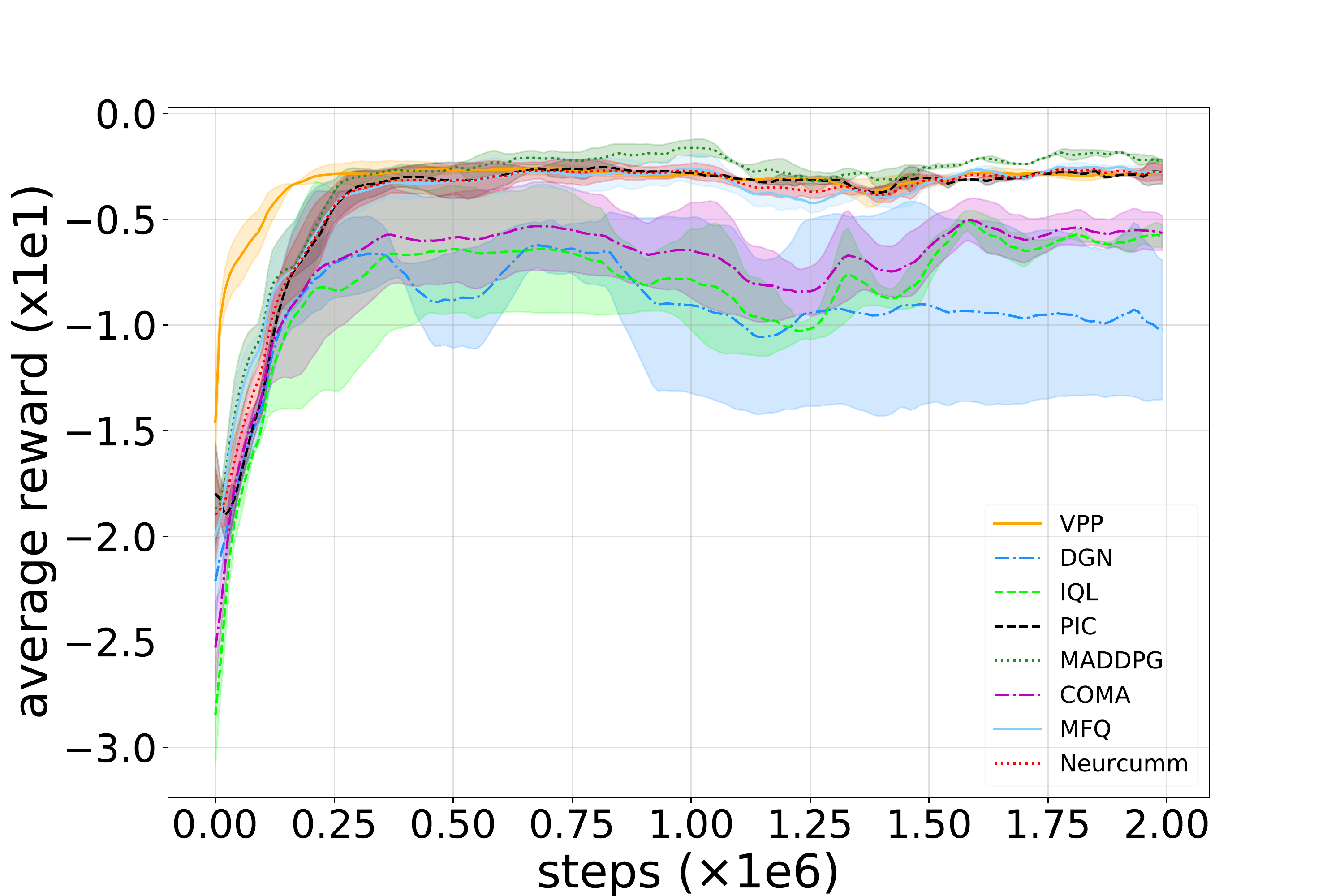}}
	\subfloat[CityFlow:7*7]{\includegraphics[width=.31\textwidth]{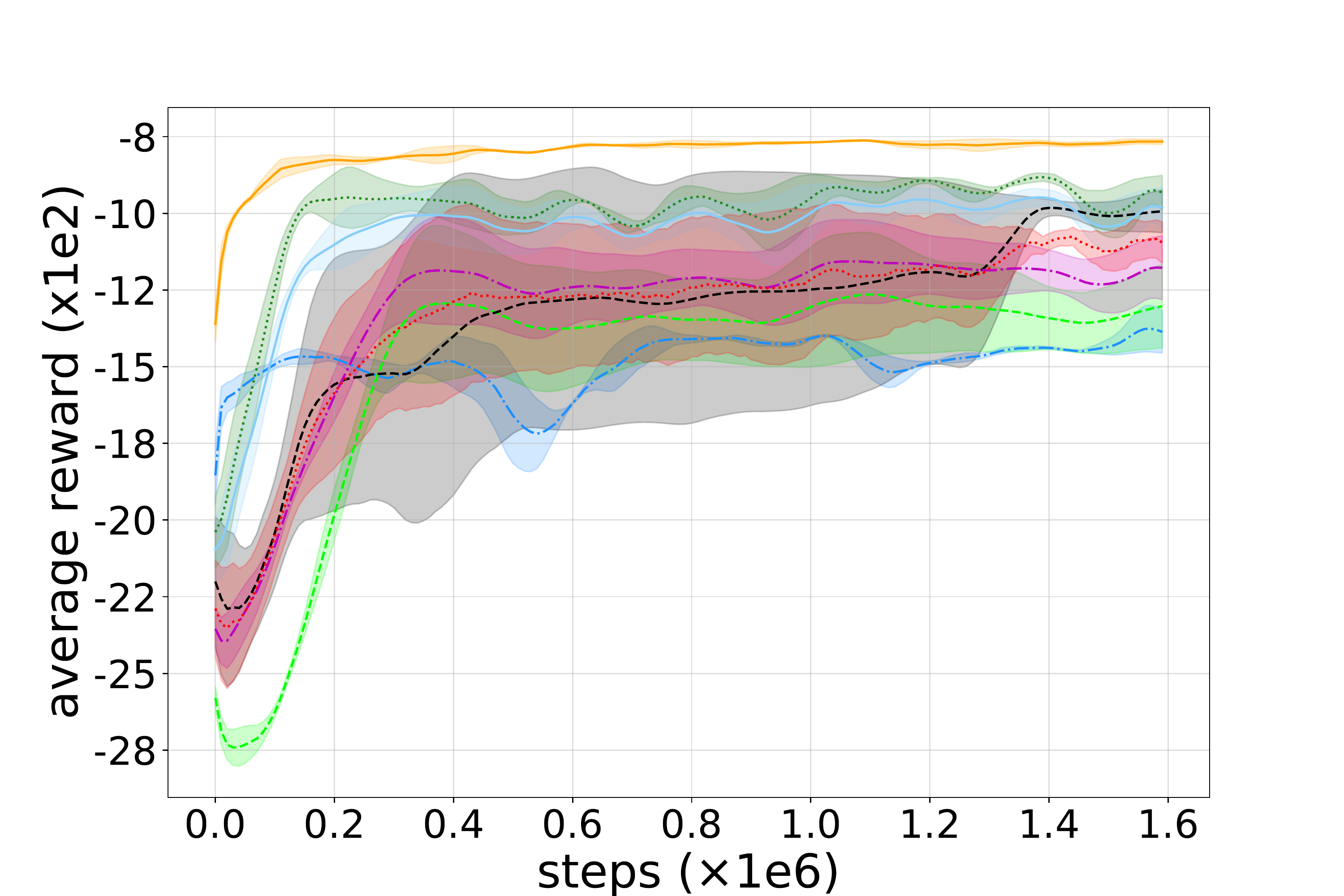}}
	\subfloat[CityFlow:15*15]{\includegraphics[width=.31\textwidth]{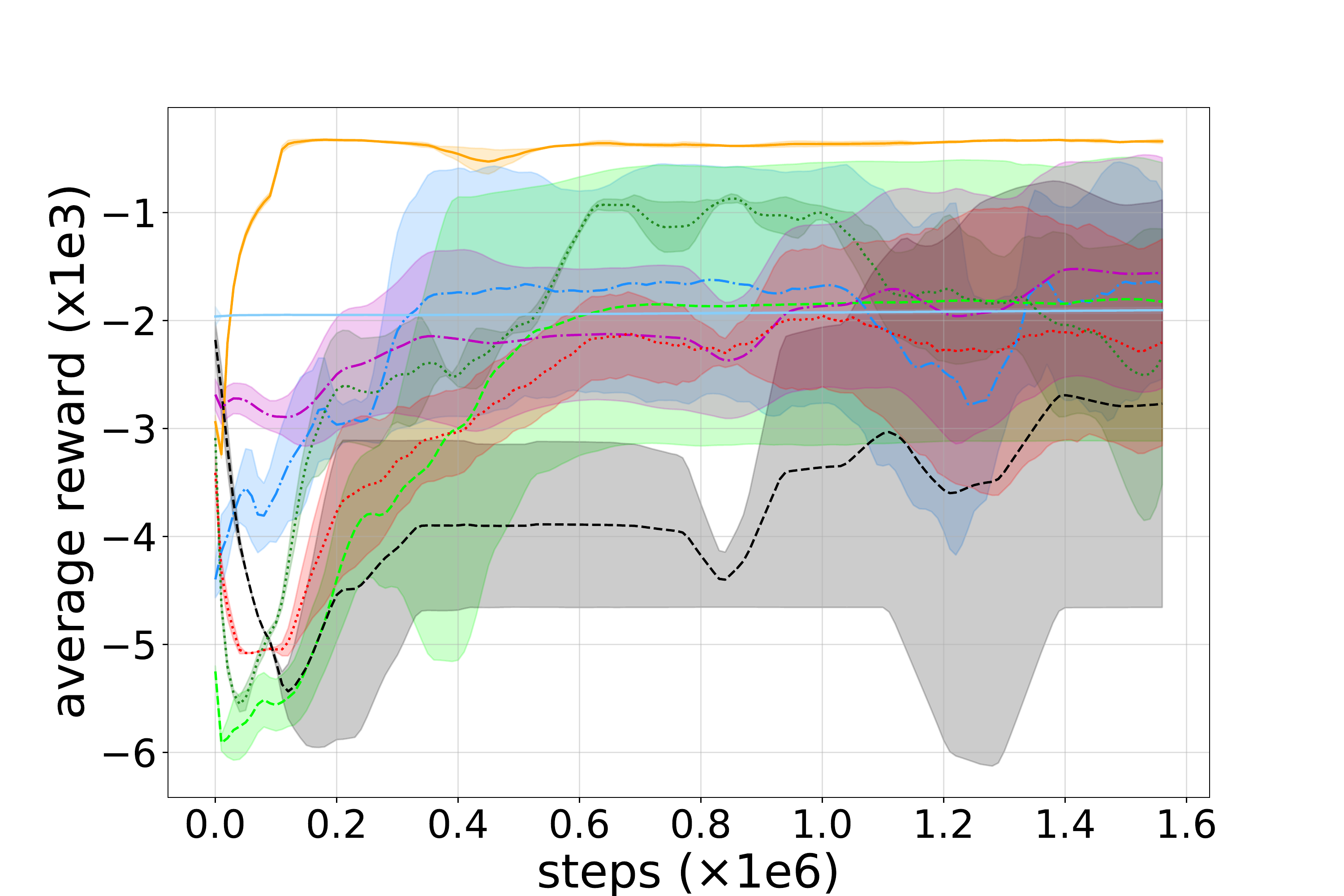}}
	\caption{Performance of different methods on traffic lights control scenarios in CityFlow environment: 
		(a) N=16 ($4\times4$ grid), Gudang sub-district, Hangzhou, China.
		(b) N=49 ($7\times7$ grid), 
		(c) N=225  ($15\times15$ grid). The horizontal axis is time steps (interaction with the environment). The vertical axis is average episode reward, which refers to negative average travel time.
		Higher rewards are better. In the small scale problem, all algorithms have similar performance. VPP obtains much better performance than all the baselines on large-scale tasks.
	}
	\label{fig:exp_futher_cityflow}
\end{figure*}

\begin{figure*}
	\centering
	\subfloat[]{\includegraphics[width=.31\textwidth]{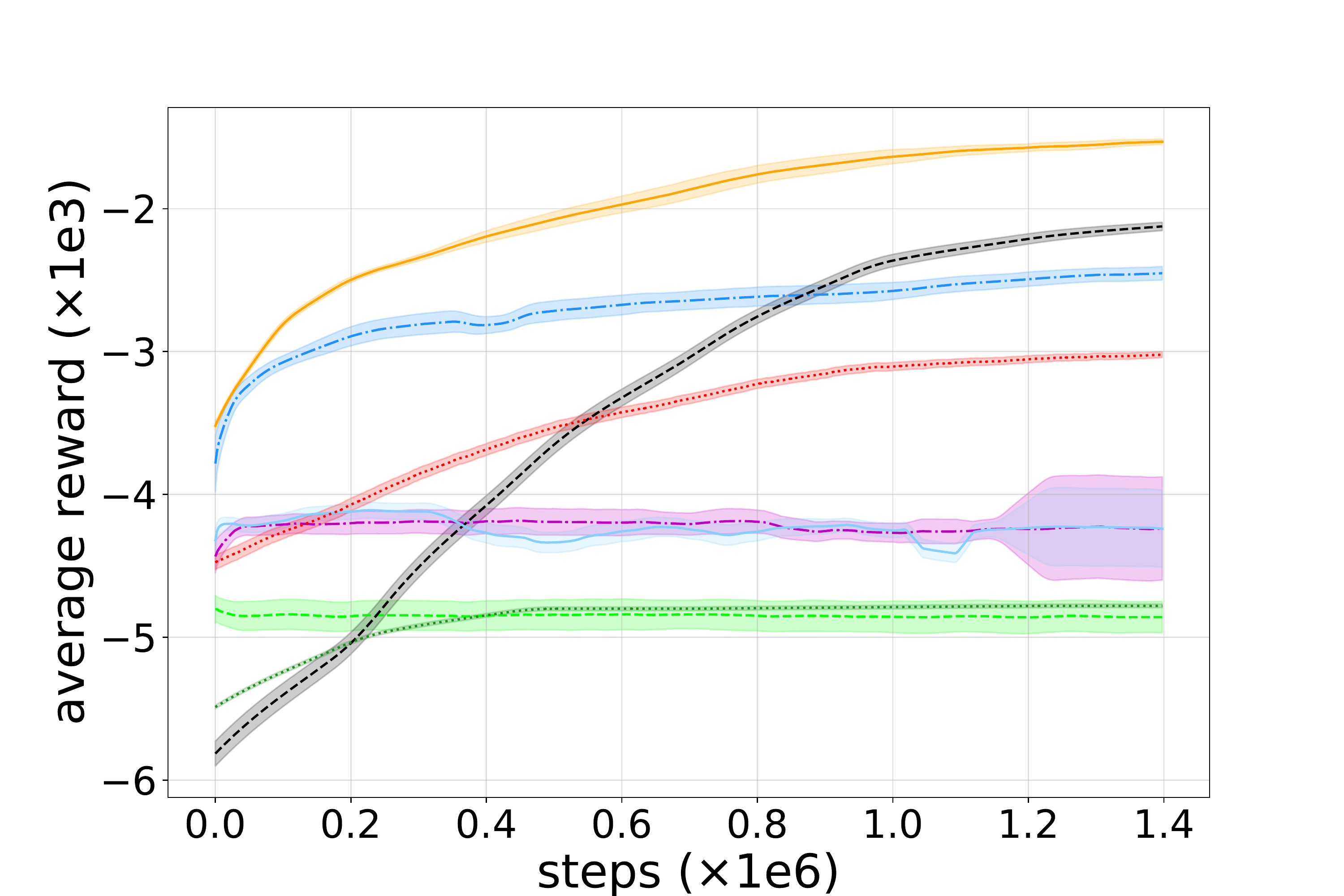}}\quad
	\subfloat[]{\includegraphics[width=.31\textwidth]{figs_new/co-navigation30.pdf}}\quad
	\subfloat[]{\includegraphics[width=.31\textwidth]{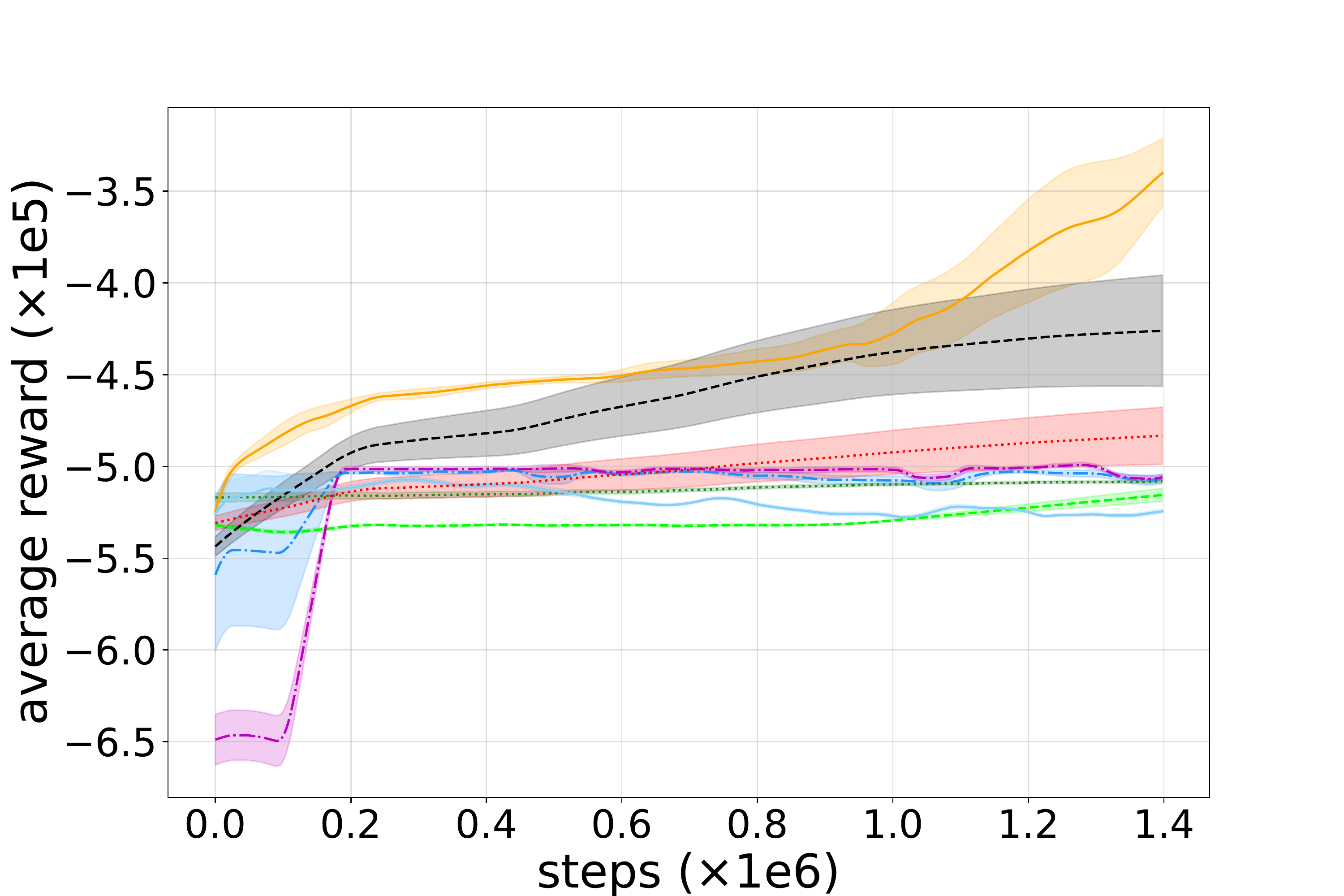}}\quad
	\subfloat[]{\includegraphics[width=.31\textwidth]{figs_new/Jungle.pdf}}\quad
	\subfloat[]{\includegraphics[width=.31\textwidth]{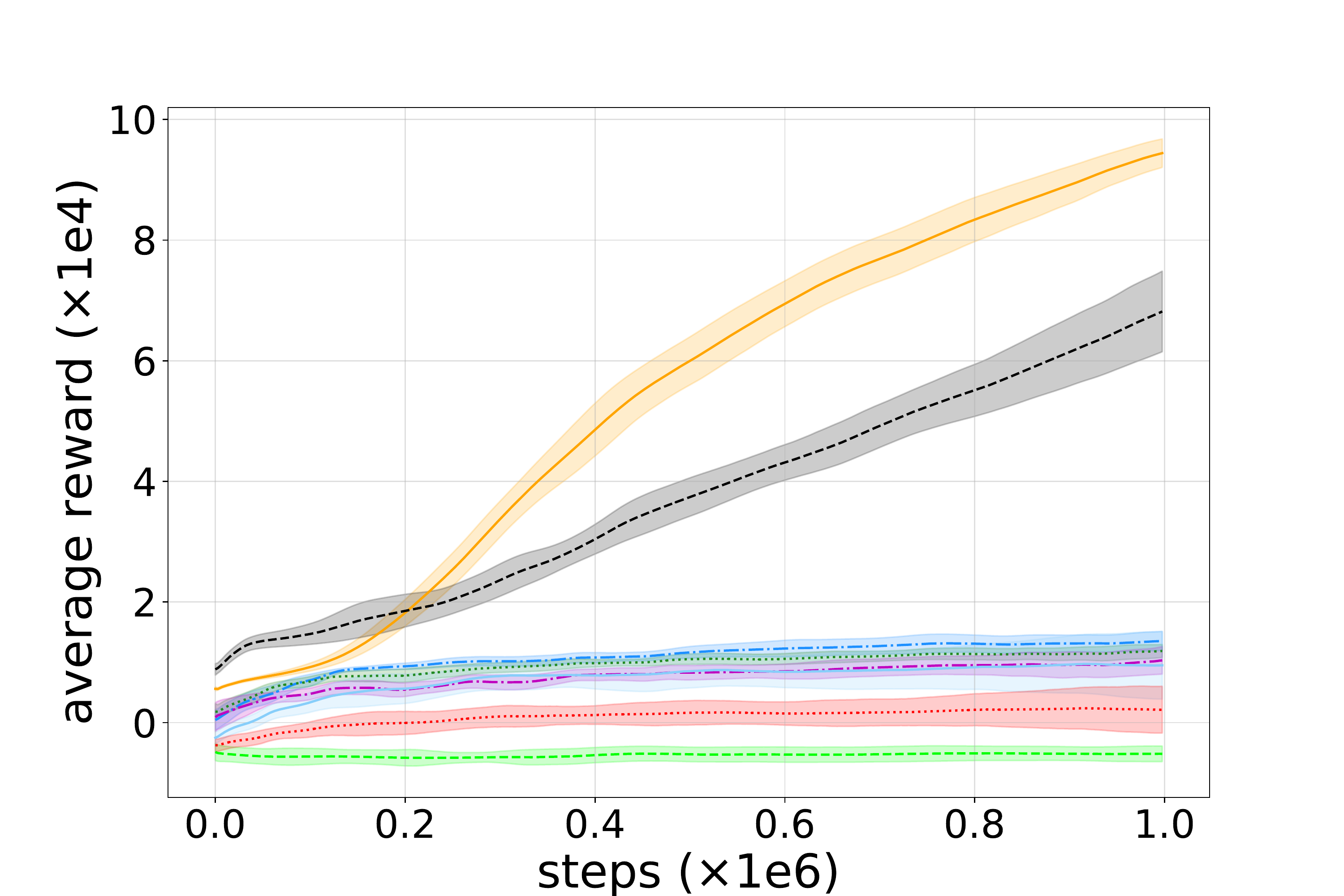}}\quad
	\subfloat[]{\includegraphics[width=.31\textwidth]{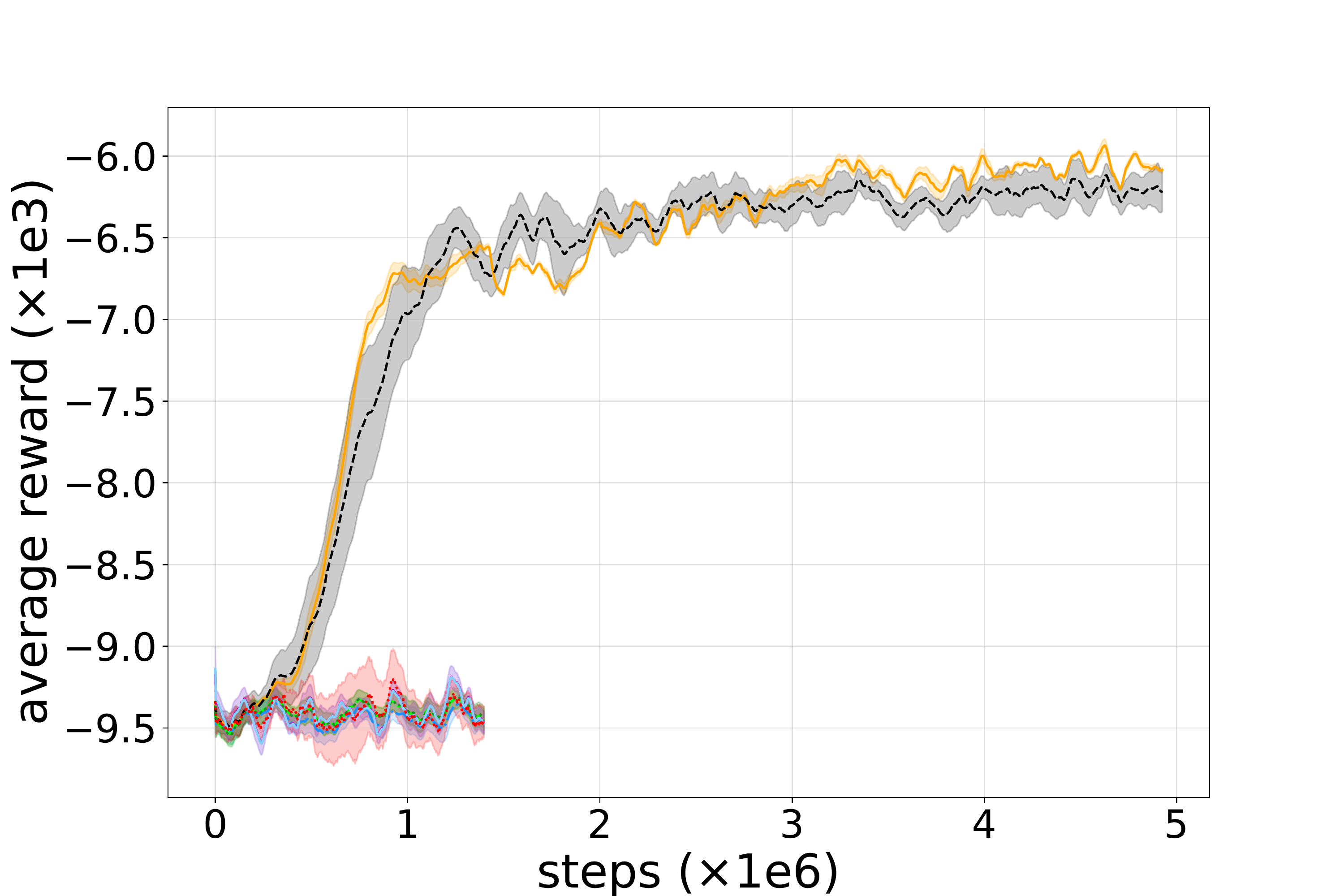}} \quad
	\caption{Comparison with baselines on different scenarios. \textit{cooperative navigation:} (a) N=15, (b) N=30 and (c) N=200 . (d) \textit{jungle} (N=20, F=12), (e) \textit{prey and predator} (N=100), and (f) \textit{cooperative push} (N=100). 
		The horizontal axis is environmental steps (number of interactions with the enviroment). The vertical axis is average episode reward.
		The larger average reward indicates better result. The proposed variational policy propagation~(VPP) beats all the baselines on different scale of instances.
	}
	\label{fig:exp_futher_scenarios}
\end{figure*}

For most of the experiments, we run them long enough with  1 million to 1.5 million steps and stop (even in some cases  our algorithm does not converge to the asymptotic result), since every  experment in MARL may cost several days. We present more results on Cityflow in Figure \ref{fig:exp_futher_cityflow}. In the small scale setting, all algorithms have similar performance while in the large scale setting VPP beats all the others.  Figure~\ref{fig:exp_futher_scenarios} (a)(b)(c) provide experimental results on the cooperative navigation instances with $N=15$, $N=30$ and $N=200$ agents. It's clear that  MADDPG performs well in the small setting (N=15), however, they failed in large-scale instances ($N=30$ and $N=200$).   DGN using graph convolutional network can hold the property of permutation invariance and improve the performance of the multi-agent system. However, it also fails to solve the large-scale settings with $N=200$ agents. Empirically, after $1.5 \times 10^6$ steps, PIC obtains a large reward ($-425085 \pm 31259$) on this large-scale setting. VPP approaches $-329229 \pm 14730$ and is much better than PIC.
Furthermore, Figure~\ref{fig:exp_futher_scenarios} shows the results of different methods on (d) jungle (N=20, F=12) and (e) \textit{prey and predator} (N=100). The experimental results show our method  can beat all baselines on these two tasks.
On the scenario of \textit{cooperative push} (N=100) as shown in Figure~\ref{fig:exp_futher_scenarios}(f),
it's clear that DGN, MADDPG and  Neurcumm  fail to converge to good rewards after $1.5 \times 10^6$ environmental steps.
In contrast, PIC and VPP obtain much better rewards than these baselines.
Limited by the computational resources, we only show the long-term performance of the best two methods. Figure~\ref{fig:exp_futher_scenarios}(f) shows that VPP is slightly better than PIC in this setting.

\section{Hyperparameters}
\label{app:hypers}

The parameter on the environment: For the max episode length, we follow the similar settings like that in the baselines \citep{lowe2017multi} . Particularly, we set 25 for MPE and set 100 for CityFlow. For MAgent, we find that setting the max episode length by 25 is better than 100.  All the methods share the same setting.

We list the range of hyperparameter that we tune in all baselines and variational policy propagation. $\gamma : \{0.95,0.98,0.99, 0.999\}$,  learning rate $:\{1,5,10,100\}\times$1e-4. activation function: $\{relu, gelu, tanh\}$, batch size:$\{128, 256, 512, 1024\}$, gradient steps: $\{1, 2, 4, 8\}$. Number of hidden units in MLP: $\{32,64,128,256,512\}$, number of layers in MLP:$\{1,2,3\}$ in all experiment. Regarding to VPP, the first MLP-layer which maps the state to $\tilde{u}_i^0$ and $\tilde{v}_{ij}^0$ has 32 hidden units. Then we use 2-head attention with dimension 32. The other MLP-layers has 128 hidden units. Other parameters of VPP are reported in ~Table.\ref{tab:shared_params}. 

\begin{table}[H]
	\renewcommand{\arraystretch}{1.1}
	\centering
	\caption{Hyperparameters}
	\label{tab:shared_params}
	\vspace{1mm}
	\begin{tabular}{|l| l| }
		\hline
		Parameter &  Value \\
		\hline
		optimizer &Adam\\
		learning rate of all networks & $0.01$\\
		discount of reward &  0.95\\
		replay buffer size & $10^6$\\
		max episode length in MAgent & 25 \\
		max episode length in MPE, CityFlow & 100 \\
		number of samples per minibatch & 1024\\
		nonlinearity & ReLU\\
		target network smoothing coefficient ($\tau$)& 0.01\\
		Entropy regularizer factor($\alpha$) & 0.2 \\
		\hline
	\end{tabular}
\end{table}

\section{Derivation }\label{section:derivation}

\subsection{Proof of proposition \ref{proposition:MRF}}\label{section:proof} 
We prove the result by induction using the backward view. 

To see that, plug  $r(s^t,\mathbf{a}^t)=\sum_{i=1}^{N} r_i(s^t,a_i^t,a_{\mathcal{N}_i}^{t})$ into the distribution of the optimal policy \cite{levine2018reinforcement} in section \ref{section:background}. 
\begin{align*}
p(\tau)=[p(s^0)\prod_{t=0}^{T}
p(s^{t+1}|s^t,\mathbf{a^t})]\exp\sum_{t=0}^T \sum_{i=1}^N r_i(s^t, a_i^t, a_{\mathcal{N}_i}^t )
\end{align*}
Recall the goal is to find the best approximation of $\pi(\mathbf{a}^t|s^t)$ such that the trajectory distribution $\hat{p}(\tau)$ induced by this policy can match the optimal trajectory probability $p(\tau)$. Thus we minimize the KL divergence between them
$\min_{\pi}  D_{KL}(\hat{p}(\tau)|| p(\tau)),$
where $\hat{p}(\tau)=p(s^0) \prod_{t=0}^{T}p(s^{t+1}|s^t, \mathbf{a^t})\pi(\mathbf{a^t}|s^t)$ \cite{levine2018reinforcement}.
We can do optimization w.r.t. $\pi(a^t|s^t)$ as that in \citep{levine2018reinforcement} and obtain a backward algorithm on the policy $\pi^*(\mathbf{a}^t|s^t)$ (See equation \ref{equ:appendix_optimal} in  \ref{app:app_derivation}.)
\begin{flalign}
\label{equ:optimal_policy}
\begin{aligned}
\pi^*(\mathbf{a^t}|s^t)= \frac{1}{Z}\exp \big(\mathbb{E}_{p(s^{t+1:T}, \mathbf{a}^{t+1:T}|s^t,\mathbf{a}^t)}[ \sum_{t'=t}^{T} \sum_{i=1}^{N} r_i(s^{t'},a_i^{t'},a_{\mathcal{N}_i}^{t'})-\sum_{t'=t+1}^T \log \pi(\mathbf{a^{t'}}|s^{t'}) ] \big).
\end{aligned}
\end{flalign}

Using the result \eqref{equ:optimal_policy}, when $t=T$, the optimal policy is 
$$ \pi^*(\mathbf{a}^T|s^T)=\frac{1}{Z}\exp(\sum_{i=1}^{N} r_i(s^T,a_i^T, a^T_{\mathcal{N}_i})).$$
Obviously, it satisfies the form 
$\pi^*(\mathbf{a}^T|s^T)=\frac{1}{Z}\exp(\sum_{i=1}^{N} \psi_i(s^T, a_i^T, a_{\mathcal{N}_i}^T)).$

Now suppose from step $t+1$ to $T$, we have

\begin{equation}\label{equ:appendix_induction1}
\pi^*(\mathbf{a}^{t'}|s^{t'})=\frac{1}{Z}\exp(\sum_{i=1}^{N} \psi_i(s^{t'}, a_i^{t'}, a_{\mathcal{N}_i}^{t'})) 
\end{equation}
for $t'= t+1,...,T $.

Recall that we have the result
\begin{flalign} \label{equ:appendix_induction2}
\begin{aligned}
\pi^*(\mathbf{a^t}|s^t)
= \frac{1}{Z}\exp \big(\mathbb{E}_{p(s^{t+1:T}, \mathbf{a}^{t+1:T}|s^t,\mathbf{a}^t)}[ 
\sum_{t'=t}^{T} \sum_{i=1}^{N} r_i(s^{t'},a_i^{t'},a_{\mathcal{N}_i}^t)-\sum_{t'=t+1}^T \log \pi^*(\mathbf{a^{t'}}|s^{t'}) ] \big).
\end{aligned}
\end{flalign}

Now plug \eqref{equ:appendix_induction1} into \eqref{equ:appendix_induction2}, we have 
\begin{flalign}
\begin{aligned}
\pi^*(\mathbf{a^t}|s^t)
= \frac{1}{Z}\exp \big(\mathbb{E}_{p(s^{t+1:T}, \mathbf{a}^{t+1:T}|s^t,\mathbf{a}^t)}[
\sum_{t'=t}^{T} \sum_{i=1}^{N} r_i(s^{t'},a_i^{t'},a_{\mathcal{N}_i}^{t'})-\sum_{t'=t+1}^T 
\sum_{i=1}^N\psi_i(s_i^{t'},a_i^{t'},a_{\mathcal{N}_i}^{t'})
+C] \big),
\end{aligned}
\end{flalign}
where $C$ is some constant related to the normalization term. Thus, we redefine a new term 
\begin{flalign}
\begin{aligned}
\Tilde{\psi}_i(s^t,a^t,a_{\mathcal{N}_i}^t)=\mathbb{E}_{p(s^{t+1:T}, a^{t+1:T}|s^t,a^t)}\big[\sum_{t=t'}^{T}\big( r_i(s^{t'},a_i^{t'},a_{\mathcal{N}_i}^{t'})-\sum_{t'=t+1}^{T} \psi_i(s^{t'},a^{t'},a_{\mathcal{N}_i}^{t'}) \big)\big].
\end{aligned}
\end{flalign}

Then obviously $\pi^{*}(\mathbf{a}^t|s^t)$ satisfies the form what we need by absorbing the constant $C$ into the normalization term . Thus we have the result.

\subsection{Derivation of the algorithm}\label{app:app_derivation}
We start the derivation with minimization of the KL divergence $KL(\hat{p}(\tau)||p(\tau))$, where  $ p(\tau)=[p(s^0)\prod_{t=0}^{T}p(s^{t+1}|s^t,\mathbf{a^t})]\exp\big(\sum_{t=0}^T \sum_{i=1}^N r_i(s^t, a_i^t, a_{\mathcal{N}_i}^t )\big)$, 
$\hat{p}(\tau)=p(s^0) \prod_{t=0}^{T}p(s^{t+1}|s^t, \mathbf{a^t})\pi(\mathbf{a^t}|s^t).$

\begin{equation}
\begin{split}
KL(\hat{p}(\tau)||p(\tau))=&\mathbb{E}_{\tau\sim \hat{p}(\tau)} \sum_{t=0}^{T}\big(  \sum_{i=1}^{N} r_i(s^t,a_i^t,a_{\mathcal{N}_i}^i)-\log \pi(\mathbf{a}^t|s^t)  \big) \\  
=&\sum_{\tau} [p(s^0) \prod_{t=0}^{T}p(s^{t+1}|s^t, \mathbf{a^t})\pi(\mathbf{a^t}|s^t)]\sum_{t=0}^{T}\big(  \sum_{i=1}^{N} r_i(s^t,a_i^t,a_{\mathcal{N}_i}^t)-\log \pi(\mathbf{a}^t|s^t)  \big).
\end{split}
\end{equation}

Now we optimize KL divergence w.r.t $\pi(\cdot|s^t)$. 
Considering the constraint $\sum_j \pi(j|s^t)=1 $, we introduce a Lagrangian multiplier $\lambda(\sum_{j=1}^{|\mathcal{A}|}\pi(j|s^t)-1)$.
Now we take gradient of $KL(\hat{p}(\tau)|| p(\tau))+\lambda(\sum_{j=1}^{|\mathcal{A}|} \pi(j|s^t)-1)$ w.r.t $\pi(\cdot|s)$, set it to zero, and obtain 

$$\log \pi^*(\mathbf{a}^t|s^t)= \mathbb{E}_{p(s^{t+1:T}, \mathbf{a}^{t+1:T}|s^t,\mathbf{a}^t)}[ 
\sum_{t'=t}^{T} \sum_{i=1}^{N} r_i(s^{t'},a_i^{t'},a_{\mathcal{N}_i}^{t'})-\sum_{t'=t+1}^T \log \pi(\mathbf{a^{t'}}|s^{t'}) ] -1+\lambda.$$

Therefore 

$$ \pi^*(\mathbf{a}^t|s^t)\propto \exp \big(\mathbb{E}_{p(s^{t+1:T}, \mathbf{a}^{t+1:T}|s^t,\mathbf{a}^t)}[ 
\sum_{t'=t}^{T} \sum_{i=1}^{N} r_i(s^{t'},a_i^{t'},a_{\mathcal{N}_i}^{t'})-\sum_{t'=t+1}^T \log \pi(\mathbf{a^{t'}}|s^{t'}) ]  \big).  $$

Since we know $\sum_{j}\pi(j|s^t)=1$, thus we have 

\begin{equation}\label{equ:appendix_optimal}
\pi^*(\mathbf{a}^t|s^t)=\frac{1}{Z}\exp \big(\mathbb{E}_{p(s^{t+1:T}, \mathbf{a}^{t+1:T}|s^t,\mathbf{a}^t)}[ 
\sum_{t'=t}^{T} \sum_{i=1}^{N} r_i(s^{t'},a_i^{t'},a_{\mathcal{N}_i}^{t'})-\sum_{t'=t+1}^T \log \pi(\mathbf{a^{t'}}|s^{t'}) ]  \big).  
\end{equation}

For convenience, we define the soft $V$ function and $Q$ function as that in \citep{levine2018reinforcement}, and will show how to decompose them into $V_i$ and $Q_i$ later.

\begin{equation}\label{equ:definition_VQ}
\begin{split}
V(s^{t+1}):=\mathbb{E} \big[\sum_{t'=t+1}^T \sum_{i=1}^{N} r_i(s^{t'},a^{t'}_i,a^{t'}_{\mathcal{N}_i}) -\log \pi(\mathbf{a^{t'}}|s^{t'})|s^{t+1}\big],\\
Q(s^t,\mathbf{a}^t)
:= \sum_{i=1}^{N} r_i(s^t,a^t_i,a_{\mathcal{N}_i}^t) + \mathbb{E}_{p(s^{t+1}|s^t,\mathbf{a}^t)}[V(s^{t+1})]      
\end{split} 
\end{equation}

Thus $V(s^t)=E_{\pi}[Q(s^t,a^t)-\log \pi(\mathbf{a^t}|s^t)]$. The optimal policy $\pi^*(\mathbf{a}^t|s^t)=\frac{\exp(Q(s^t,\mathbf{a}^t)}{\int \exp Q(s^t,\mathbf{a}^t) d\mathbf{a}^t }$ by plugging the definition of $Q$ into \eqref{equ:appendix_optimal}.


Remind in section \ref{section:reduce_policy_space}, we have approximated the optimal joint policy by the mean field approximation $\prod_{i=1}^{N}q_i(a_i|s).$ We now plug this into the definition of \eqref{equ:definition_VQ} and consider the discount factor. Notice it is easy to incorporate the discount factor by defining a absorbing state where each transition have $(1-\gamma)$ probability to go to that state. Thus we have
\begin{flalign}
\begin{aligned}
V(s^{t+1}):=\mathbb{E} \big[  \sum_{t'=t+1}^T ( \sum_{i=1}^{N} r_i(s^{t'},a^{t'}_i,a^{t'}_{\mathcal{N}_i}) -\sum_{i=1}^{N}\log q_i(a_i^{t'}|s^{t'}) )|s^{t+1}\big],\\
Q(s^t,\mathbf{a}^t)
:= \sum_{i=1}^{N} r_i(s^t,a^t_i,a_{\mathcal{N}_i}^t) + \gamma \mathbb{E}_{p(s^{t+1}|s^t,\mathbf{a}^t)}[V(s^{t+1})].
\end{aligned}
\end{flalign}

Thus we can further decompose $V$ and $Q$ into $V_i$ and $Q_i$. We define $V_i$ and $Q_i$ in the following way.
$$ V_i(s^{t+1})=\mathbb{E} [ \sum_{t'=t+1}^T \big(r_i(s^{t'},a^{t'}_i,a^{t'}_{\mathcal{N}_i})-\log q_i(a_i^{t'}|s^{t'})\big)|s^{t+1}], $$
$$ Q_i(s^t,a_i^t,a^t_{\mathcal{N}_i})
=  r_i(s^t,a^t_i,a_{\mathcal{N}_i}^t)+ \gamma \mathbb{E}_{p(s^{t+1}|s^t,\mathbf{a}^t)}[V_i(s^{t+1})]. $$

Obviously we have $V=\sum_{i=1}^N V_i$ and $Q=\sum_{i=1}^N Q_i$. 

For $V_i$, according to our definition, we obtain
\begin{equation}
V_i(s^t)=\mathbb{E}_{\mathbf{a}^t\sim \prod_{i=1}^{N} q_i } [r_i(s^t,a^t_i,a_{\mathcal{N}_i}^t)-\log q_i(a_i^t|s^t)+\mathbb{E}_{p(s^{t+1}|s^t,\mathbf{a}^t) }V_i(s^{t+1})].
\end{equation}

Now we relate it to  $Q_i$, and have

$$V_i(s^t)=\mathbb{E}_{\mathbf{a}^t\sim\prod_{i=1}^{N} q_i} [Q_i(s_i^t,a_i^t,a_{\mathcal{N}_i}^t)-\log q_i(a_i^t|s^t)]=\mathbb{E}_{(a_i,a_{\mathcal{N}_i)}\sim (q_i,q _{\mathcal{N}_i}) } Q_i(s_i^t,a_i^t,a_{\mathcal{N}_i}^t)-\mathbb{E}_{a_i\sim q_i}\log q_i(a_i^t|s^t).$$

Thus it suggests that we should construct the loss function on $V_i $ and $Q_i$ in the following way. In the following, we use parametric family (e.g. neural network) characterized by $\eta_i$ and $\kappa_i$ to approximate  $V_i$ and $Q_i$ respectively.

\begin{align*}
J(\eta_i)=\mathbb{E}_{s^t\sim D}[\frac{1}{2}\big(&V_{\eta_i}(s^t)-
\mathbb{E}_{(a_i,a_{\mathcal{N}_i)}\sim (q_i,q _{\mathcal{N}_i})}[Q_{\kappa_i}(s^t,a_i^t,a^t_{\mathcal{N}_i} )]-\log q_{i}(a_i^t|s^t)  \big)^2 ], 
\end{align*}
\begin{equation}
J(\kappa_i)=\mathbb{E}_{(s^t,a_i^t,a_{\mathcal{N}_i^t})\sim D} [\frac{1}{2}\big( Q^i_{\kappa_i}(s^t,a^i_t, a_{\mathcal{N}_i}^t)-\hat{Q}(s^t, a^i_t,a_{\mathcal{N}_i}^t) \big)^2].
\end{equation}
where $\hat{Q}_i(s^t,a_i^t, a^t_{\mathcal{N}_i})=r_i+\gamma \mathbb{E}_{s^{t+1}\sim p(s^{t+1}|s^t,a^t)}[V_{\eta_i}(s^{t+1})].$

Now we are ready to derive the update rule of the policy, i.e., the variational policy propagation network.

Remind the variational policy propagation network actually is a mean-field approximation of the joint-policy.
$$ \min_{p_1,p_2,...,p_n} KL(\prod_{i=1}^{N} p_i(a_i|s)|| \pi(\mathbf{a}|s) ).$$
It is the optimization over the \emph{function} $p_i$ rather than certain parameters. We have proved that  after $M$ iteration of variational policy propagation, we have output the nearly optimal solution $q_i$.

In the following, we will demonstrate how to update the parameter $\theta$ of the policy. Again we minimize the KL divergence
$$ \min_\theta \mathbb{E}_{s^t} KL( \prod_{i=1}^{N} q_{i,\theta} (a^t_i|s^t) || \pi(\mathbf{a}^t|s^t)  )  $$

Plug the $\pi(\mathbf{a}^t|s^t)=\frac{\exp(Q(s^t,\mathbf{a}^t))}{\int \exp Q(s^t,\mathbf{a}^t) d\mathbf{a}_t }$ into the KL divergence. It is easy to see, it is equivalent to the following the optimization problem by the definition of the KL divergence. 
$$\max_{\theta}\mathbb{E}_{s^t}\big[\mathbb{E}_{\mathbf{a}^t\sim \prod q_{i,\theta}(a_i^t|s^t)}[\sum_{i=1}^{N}Q_{\kappa_i}(s^t,a_i^t,a_{\mathcal{N}_i}^t)-\sum_{i=1}^N\log q_{i,\theta}(a_i^t|s^t)] \big].$$
Thus we sample state from the replay buffer and have the loss of the policy as

$$ J(\theta)= \mathbb{E}_{s^t\sim D, \mathbf{a^t}\sim \prod_{i=1}^N q_{i,\theta}(a_i^t|s^t)} [\sum_{i=1}^{N}\log q_{i,\theta}(a^t_i|s^t)-\sum_{i=1}^{N} Q_{\kappa_i}(s^t, a_i^t, a_{\mathcal{N}_i}^t) ].$$

\end{document}